\documentclass{article}

\usepackage[eandd, final]{neurips_2026}

\usepackage[utf8]{inputenc} % allow utf-8 input
\usepackage[T1]{fontenc}    % use 8-bit T1 fonts
\usepackage[colorlinks=true, citecolor=blue, linkcolor=blue, urlcolor=blue]{hyperref}       % hyperlinks
\usepackage{url}            % simple URL typesetting
\usepackage{booktabs}       % professional-quality tables
\usepackage{amsfonts}       % blackboard math symbols
\usepackage{nicefrac}       % compact symbols for 1/2, etc.
\usepackage{microtype}      % microtypography
\usepackage{xcolor}         % colors
\usepackage{tikz}
\usepackage{pgfplots}
\usepackage{siunitx}
\usepackage{multirow}
\usepackage{wrapfig}
\usepackage{amssymb}
\usepackage{subcaption}
\usepackage{graphicx, siunitx, caption}
\usepackage{listings}
\lstdefinestyle{appendixcode}{%
  basicstyle=\ttfamily\scriptsize,
  breaklines=true,
  breakatwhitespace=false,
  columns=fullflexible,
  keepspaces=true,
  showstringspaces=false
}
\lstdefinestyle{appendixtext}{%
  basicstyle=\ttfamily\tiny,
  breaklines=true,
  breakatwhitespace=false,
  columns=fullflexible,
  keepspaces=true,
  showstringspaces=false
}
\usepgfplotslibrary{groupplots}

\usetikzlibrary{arrows.meta, positioning, fit, calc, backgrounds}

\pgfplotsset{compat=1.18}

\definecolor{slsblue}{RGB}{56,94,173}
\definecolor{slsgreen}{RGB}{45,140,110}
\definecolor{slsorange}{RGB}{214,133,27}
\definecolor{slsred}{RGB}{180,72,72}

\definecolor{questionfill}{RGB}{216,223,234} % soft slate blue
\definecolor{logfill}{RGB}{218,228,220}      % muted sage
\definecolor{answerfill}{RGB}{233,229,207}   % warm stone (replaces yellow)
\definecolor{scorefill}{RGB}{232,222,226}    % dusty mauve
\definecolor{answerline}{RGB}{145,120,88}    % muted bronze/taupe
\definecolor{bordergray}{RGB}{82,82,82}
\definecolor{outergray}{RGB}{184,184,184}

\definecolor{openaiblue}{RGB}{88,107,180}
\definecolor{anthropicorange}{RGB}{176,126,84}
\definecolor{modelgray}{RGB}{120,124,128}

\definecolor{accentgreen}{RGB}{138,168,84}
\definecolor{accentteal}{RGB}{78,130,136}

\definecolor{loopbg}{HTML}{F7F5F9}
\definecolor{loopborder}{HTML}{756B84}
\definecolor{bnodefill}{HTML}{ECE8F1}
\definecolor{bnodeborder}{HTML}{665D74}
\definecolor{bfeedback}{HTML}{7C748A}

\definecolor{svcpeach}{HTML}{F0D9C4}   % Ingress — warm peach
\definecolor{svcmauve}{HTML}{D4B8CB}   % Backend — dusty mauve
\definecolor{svcolive}{HTML}{C8D1A8}   % Control — soft olive
\definecolor{svcrose}{HTML}{E0BFC6}    % Backend — muted rose

\definecolor{svcblue}{HTML}{CADCF0}      % Ingress — cool blue
\definecolor{svclavender}{HTML}{D9D0E8}   % Backend — soft lavender
\definecolor{svcsage}{HTML}{CCE1CB}       % Control — muted sage green

\definecolor{cornflower}{HTML}{5B9BD5}
\definecolor{terracotta}{HTML}{e5b3a6}
\definecolor{jade}{HTML}{6BBBAA}

\definecolor{othergray}{RGB}{107,114,128}

\newcommand*{\slsbench}{\textsc{SLS-Bench}}

\title{\slsbench: A Benchmark for Incident Log Summarization with Synthetic Observability Data}

\author{
  Valerii Iakovlev \quad Markus Spanring \quad Martin Flechl\\
  Dynatrace Research\\
  \texttt{\{valerii.iakovlev, markus.spanring, martin.flechl\}@dynatrace.com}
}

\begin{document}

\maketitle

\begin{abstract}
% As IT systems grow in complexity, the need for automated, agent-driven incident response becomes increasingly critical. A key capability for these agents is summarizing large log streams into human-readable reports. Further progress in this area requires benchmarks to meaningfully evaluate this function, yet existing benchmarks leave end-to-end summarization from large log streams underexplored due to the prohibitive cost and complexity of collecting real-world data. We introduce \slsbench, a benchmark for incident log summarization featuring 90 tasks grounded in real-world incident reports, each with a synthetic log stream and a reference summary. To construct \slsbench, we developed a data generation pipeline that converts incident reports into structured incident models, which in turn guide the generation of synthetic logs and reference summaries. This pipeline combines narrative grounding, explicit incident modeling, generator--verifier loops, and statistical comparisons with real-world log streams to produce realistic and controllable observability data. Our evaluation of 14 LLMs on \slsbench\ shows that while frontier models can produce useful summaries, they do so at a substantial cost and latency, and open-weight models lag significantly. \slsbench\ provides a robust framework for the systematic evaluation of log analysis agents and a scalable blueprint for future synthetic observability benchmarks.
As IT systems grow in complexity, automated agent-driven incident response becomes increasingly critical. A key capability for these agents is summarizing large log streams into human-readable reports. Further progress requires meaningfully evaluating this function, yet existing benchmarks leave end-to-end summarization from large log streams underexplored due to the prohibitive cost and complexity of collecting real-world data. We introduce \slsbench, a benchmark for incident log summarization featuring 90 tasks grounded in real-world incident reports, each with a synthetic log stream and a reference summary. To construct \slsbench, we developed a data generation pipeline that converts incident reports into structured incident models, which guide the generation of synthetic logs and reference summaries. This pipeline combines narrative grounding, explicit incident modeling, generator--verifier loops, and statistical comparisons with real-world log streams to produce realistic and controllable observability data. Our evaluation of 14 LLMs on \slsbench\ shows that while frontier models can produce useful summaries, they do so at substantial cost and latency, and open-weight models lag significantly. \slsbench\ provides a framework for systematic evaluation of log analysis agents and a scalable blueprint for future synthetic observability benchmarks.
\end{abstract}

\section{Introduction}

% \begingroup
% \renewcommand\thefootnote{}
% \footnotetext{We open-source our code \href{https://anonymous.4open.science/r/sls-bench-110D/}{here} and data  \href{https://huggingface.co/datasets/anon-sls-bench/SLS-Bench}{here}.}
% \addtocounter{footnote}{-1}
% \endgroup

\begingroup
\renewcommand\thefootnote{}
\footnotetext{Code: \href{https://github.com/dynatrace-research/sls-bench}{github.com/dynatrace-research/sls-bench} \quad Data: \href{https://huggingface.co/datasets/dynatrace-oss/SLS-Bench}{huggingface.co/datasets/dynatrace-oss/SLS-Bench}}
\addtocounter{footnote}{-1}
\endgroup

Modern IT systems continuously generate large volumes of logs, which become essential diagnostic assets during system incidents. Diagnosing an incident requires operators to sift through vast amounts of log data, identify relevant entries, reconstruct the failure trajectory across distributed components, and distill their findings into a coherent incident summary. 
% of the failure, its root cause, and its progression.
This is one of the most impactful yet least automated log analysis tasks: it remains tedious, time-consuming, and heavily reliant on domain expertise. Agentic workflows powered by large language models (LLMs) promise to streamline this process by navigating logs, surfacing the relevant entries, and drafting summaries. Operators can then verify and refine these summaries, enabling faster incident resolution. However, whether current agents can analyze incident logs and produce comprehensive high-quality summaries remains an important open question that existing benchmarks are not fully equipped to answer.

A benchmark for incident log summarization should reflect conditions that operators face in practice: large log volumes, diverse systems and failure types, realistic failure scenarios, and tasks that require comprehensive, structured incident summaries. To our knowledge, no existing benchmark covers all four properties, leaving this end-to-end task largely unevaluated. Prior benchmarks simplify at least one dimension, for example by using short log snippets, narrow task formulations, or keyword-style outputs, rather than requiring an agent to identify a failure, trace its evolution across components, and produce a summary an operator would write after analyzing an incident. This gap persists because collecting such data at scale is prohibitively expensive, requiring access to diverse production systems, observed failures, relevant logs, and substantial expert effort to produce reference summaries.

In this work, we address this gap with a report-grounded synthetic benchmark. Rather than collecting expensive production logs and human-written summaries, we generate them synthetically by grounding them in real-world incident reports. Hundreds of such reports are publicly available, and they typically describe the system components involved, the system topology, the technology stack, the failure symptoms, the timeline of the incident, and the root cause. These are some of the core ingredients needed to construct a log stream that reflects a real system failure.

Turning these reports into benchmark tasks is non-trivial. Two main challenges exist: (1) translating high-level incident reports into detailed log streams that capture both overall incident dynamics and low-level realism (e.g., log structure and distribution), and (2) generating comprehensive reference summaries that faithfully reflect the log stream, from the overarching incident narrative down to specific log-level evidence.
\input{figures/1-pool-figure}

We address these challenges and make the following contributions:
\begin{itemize}
\item \textbf{\slsbench.} We introduce the Synthetic Log Summarization benchmark (\slsbench), consisting of 90 tasks based on curated incident reports. Each task contains a large log file (56k logs on average), a fixed question, and a comprehensive reference summary (Fig.~\ref{fig:slsbench_overview}), and is grounded in a unique incident report covering a distinct system and incident. To our knowledge, \slsbench\ is the first benchmark for incident log summarization that combines large log streams, diverse systems and incidents, structured reference summaries, and detailed multi-dimensional evaluation.
% \item \textbf{\slsbench.} We introduce the Synthetic Log Summarization benchmark (\slsbench), consisting of 90 problems from curated incident reports. Each one contains a large log file (56k logs on average) and, based on a question that is fixed across problems, a corresponding comprehensive reference summary (Fig.~\ref{fig:slsbench_overview}). Each problem is grounded in a unique incident report, covering a distinct system and incident. To our knowledge, \slsbench\ is the first benchmark for incident log summarization that combines large log streams, diverse systems and incidents, structured reference summaries, and detailed multi-dimensional evaluation.
\item \textbf{Synthetic data pipeline.} At the core of \slsbench\ is our novel report-grounded pipeline that converts real-world incident reports into synthetic logs and reference summaries, enabling benchmark construction at this scale. To ensure high quality of the generated data, we use incident models as guides for realism and validity, generator--verifier loops with strict algorithmic and rubric-based verification, and measures to ensure verifier stability. Furthermore, we demonstrate that our generated logs capture key statistical and structural properties of real-world log streams.
\item \textbf{Evaluation.} We evaluate 14 open-weight and proprietary LLMs on \slsbench\ and find that frontier models tend to achieve strong summary quality but at significant cost and latency, while open-weight models lag substantially in quality. Score breakdowns show that top-scoring models are good at high-level incident reconstruction, but weaker at capturing detailed incident progression and supporting claims with evidence. We further analyze frontier-model trajectories and identify opportunities to reduce cost and latency, including artifact pre-computation and task parallelization.
\end{itemize}

\section{Related Work}

\paragraph{Task-level log analytics.}
Automated log analysis has traditionally been organized around specialized subtasks. For instance, log parsers convert raw messages into structured templates \citep{he2017drain,dai2020logram,huo2021semparser,liu2022uniparser}; anomaly- and failure-detection methods identify unusual patterns within log sequences \citep{du2017deeplog,meng2019loganomaly,zhang2019robust,guo2021logbert}; and root-cause analysis (RCA) methods localize faulty components or connections in distributed systems \citep{chen2014causeinfer,ma2020automap,wu2020microrca,lin2024root}. While these are essential building blocks, their outputs remain highly granular: a parser returns a template, a detector returns an anomaly score, and an RCA system returns a list of candidate faults. They do not produce the comprehensive, human-readable summaries that we focus on in this work.

\paragraph{LLMs for log analysis.}
LLMs have significantly expanded the scope of tasks a single method can handle. In-context learning has been successfully applied to log parsing and anomaly detection \citep{Le2023log,jiang2023lilac,xu2024divlog,wu2024log,liu2023interpretable,qi2023loggpt}, while other research fine-tunes LLMs on log data to address multiple analytics tasks simultaneously \citep{huang2023logqa,ji2024adapting,liu2024loglm}. More recent methodologies incorporate expert-like analysis workflows and stepwise reasoning \citep{ma2025log}. Despite these advancements, evaluation protocols continue to center on limited contexts and brief outputs, falling short of full incident reconstruction and summarization.

\paragraph{Agentic log analysis.}
Agentic systems expand upon these foundations by managing diagnostic workflows end-to-end. For example, RCAgent \citep{wang2023rcaagent} predicts root causes, evidence, and solutions for cloud incidents; Flow-of-Action \citep{pei2025flow} executes multi-step RCA; AetherLog \citep{cui2025aetherlog} performs log-based RCA utilizing knowledge graphs. Additionally, systems like TrioXpert \citep{sun2025trioxpert} and COCA \citep{li2025coca} handle anomaly triage, localize faulty components, and generate brief root-cause summaries from issue reports. Yet, the evaluation of these systems remains constrained to localized components and short explanations, lacking the scale required to reconstruct complete post-incident reports from raw log streams.

\paragraph{Log analytics benchmarks.}
While existing benchmarks have grown in both scale and complexity, their evaluation targets remain heavily constrained. Datasets like Loghub-2.0 \citep{jiang2023large} provide millions of log lines but restrict tasks to template prediction. Others, such as LogQA \citep{huang2023logqa} and NLPLog \citep{ji2024adapting}, pose questions answerable from a single log entry; LogSummary \citep{meng2023log} and LogEval \citep{cui2024logeval} summarize windows of up to twenty logs into keywords or triples; and LogLM \citep{liu2024loglm} spans multiple log analysis tasks with at most ten log entries per instance. Recent large-scale telemetry benchmarks, including RCAEval \citep{pham2025rcaeval} and OpenRCA \citep{xu2025openrca}, span metrics, logs, and traces, scoring agents on top-k predictions of the root-cause service or a handful of categorical fields like failure type. Even environments featuring live fault-injection, such as AIOpsLab \citep{chen2025aiops} and Cloud-OpsBench \citep{wang2026cloudopsbench}, add deterministic state snapshots and score agents per-task on detection, localization, or mitigation success. In every case, evaluation targets small, task-specific outputs. None demand the comprehensive, stream-level failure summaries that \slsbench\ requires.

\section{Benchmark}
\label{sec:benchmark}
% Each \slsbench\ problem comprises a log file, a question requesting a structured incident summary, and a reference answer (Fig.~\ref{fig:slsbench_overview}).
% This section defines the benchmark task and details the pipeline used to generate them (Fig.~\ref{fig:slsbench_generation}). The pipeline grounds each problem in a real-world incident report, converts the report into a unified representation (\emph{incident model}), and then generates both the synthetic log stream and the reference answer. The cost to construct \slsbench\ is outlined in  App.~\ref{app:dataset_constr_cost}.

Each \slsbench\ problem consists of a question, a log file, and a reference answer (Fig.~\ref{fig:slsbench_overview}a). The question is fixed across all problems, while the log file and reference answer are produced from a unique incident report for each problem. The question asks an agent to analyze the log file and produce an incident summary. The summary must include the inferred system structure, a phase-by-phase walkthrough of system behaviors, and a diagnosis with root cause, failure chain, evidence, and confidence. The produced answer is scored against the reference answer by an LLM judge. This section describes the pipeline used to construct the tasks: Sec.~\ref{sec:components} describes the main components, and Sec.~\ref{sec:generation} describes how they are generated. Sec.~\ref{sec:scoring} describes the LLM judge. Throughout this section, we use LLMs to assist with report collection and filtering, generate and verify incident models, simulator code, and reference answers, and create judge calibration data.

\subsection{Pipeline Components}
\label{sec:components}

\paragraph{Incident reports.}

Incident reports, written by engineers after system failures, provide the real-world grounding for our pipeline. High-quality reports typically describe the affected system components, their interactions, the technology stack, observed symptoms, the failure timeline, and the root cause. Using these reports helps our synthetic data cover diverse and realistic system architectures, failure types, and incident progressions. We collected 343 publicly available incident reports through manual and LLM-assisted web search and human screening, then de-duplicated them using Jaccard similarity (threshold $0.8$), yielding 289 unique reports. We then applied an LLM-based filter designed to select reports with rich real-world grounding, such as clear descriptions of the system, failure, timeline, and symptoms. This filtering process isolated 90 high-quality reports that seed our benchmark problems. The collection process, filtering prompt, report distributions, and licensing details are provided in App.~\ref{app:filtering-prompt}.

To validate the LLM filter, we manually annotated 56 of the 289 unique reports (25 high-quality, 31 low-quality) and evaluated models from the GPT-5.2 family on filtering precision. GPT-5.2 (medium) achieved the highest precision ($91\%$) and was therefore selected as our filtering model. We assessed the filter's reliability by running it twice across all 289 reports, resulting in 266 identical verdicts (Cohen's $\kappa = 0.81$), indicating strong consistency~\citep{mchugh2012inter}.

\begin{figure*}[tb]
\centering
\resizebox{\linewidth}{!}{%
\begin{tikzpicture}[
    x=1cm,
    y=1cm,
    font=\small,
    box/.style={
        draw=bordergray!85,
        line width=0.42pt,
        rounded corners=1.9pt,
        inner sep=4.2pt,
        minimum height=1.02cm,
        align=center
    },
    givenbox/.style={
        box,
        draw=bordergray!85
    },
    llmbox/.style={
        box,
        draw=accentteal!18!black,
        line width=0.92pt
    },
    detbox/.style={
        box,
        draw=bordergray!85,
        double=white,
        double distance=0.90pt
    },
    source/.style={
        givenbox,
        fill=black!10,
        minimum width=1.8cm,
        font=\bfseries\normalsize
    },
    incmod/.style={
        llmbox,
        fill=gray!20,
        minimum width=1.8cm,
        font=\bfseries\normalsize
    },
    incsim/.style={
        llmbox,
        fill=logfill!52,
        minimum width=2.2cm,
        font=\bfseries\normalsize
    },
    outq/.style={
        detbox,
        fill=questionfill!58,
        minimum width=1.9cm,
        font=\bfseries\normalsize
    },
    outlogs/.style={
        detbox,
        fill=logfill!52,
        minimum width=1.9cm,
        font=\bfseries\normalsize
    },
    outans/.style={
        llmbox,
        fill=answerfill!58,
        minimum width=2.18cm,
        font=\bfseries\normalsize
    },
    plain/.style={
        box,
        fill=bnodefill!90,
        draw=bnodeborder,
        line width=0.78pt,
        minimum width=1.95cm,
        font=\bfseries\normalsize
    },
    flow/.style={
        -{Latex[length=2.6mm,width=2.0mm]},
        draw=black!72,
        line width=0.92pt
    },
    detflow/.style={
        -{Latex[length=2.45mm,width=1.85mm]},
        draw=bordergray!78,
        line width=0.80pt
    },
    loopframe/.style={
        fill=loopbg!50,
        draw=loopborder,
        line width=0.88pt,
        rounded corners=2.5pt
    },
    paneltag/.style={font=\bfseries\small, text=black!72},
    feedbacklabel/.style={
        font=\footnotesize,
        text=bfeedback
    },
    feedbackflow/.style={
        -{Latex[length=2.45mm,width=1.85mm]},
        draw=bfeedback,
        line width=0.82pt
    }
]

\def\leftgap{0.48}
\def\outgap{0.55}
\def\pairgap{0.40}
\def\ygap{0.78}

% Shared vertical extent for both panels.
\path[use as bounding box] (-0.12,-1.28) rectangle (15.85,1.28);

% =========================
% Panel a
% =========================
\node[source, anchor=west, dashed] (report) at (0.00,0.00)
    {Incident\\[-0.15em]report};

\node[incmod, anchor=west] (model)
    at ([xshift=\leftgap cm]report.east)
    {Incident\\[-0.15em]model};

\node[outans, anchor=west] (refa)
    at ([xshift=\outgap cm,yshift=\ygap cm]model.east)
    {Reference\\[-0.15em]answer};

\node[outq, anchor=west] (question)
    at ([xshift=\pairgap cm]refa.east)
    {Question};

\node[incsim, anchor=west] (sim)
    at ([xshift=\outgap cm,yshift=-\ygap cm]model.east)
    {Incident\\[-0.15em]simulator};

\node[outlogs, anchor=west] (logs)
    at ([xshift=\pairgap cm]sim.east)
    {Logs};

\coordinate (split) at ($(model.east)+(0.20,0)$);

\draw[flow] (report.east) -- (model.west);
\draw[flow] (model.east) -- (split) |- (refa.west);
\draw[flow] (model.east) -- (split) |- (sim.west);
\draw[flow] (sim.east) -- (logs.west);

\node[font=\bfseries\Large, text=black!68]
    at ($(refa.east)!0.5!(question.west)$) {+};

\node[paneltag, anchor=south west]
    at ([xshift=-1mm,yshift=6.5mm]report.north west) {a)};

% Small legend for panel a (bottom-left, stacked)
\coordinate (legA) at ([xshift=0.04cm,yshift=-0.49cm]report.south west);

\node[llmbox, fill=white, minimum width=0.34cm, minimum height=0.20cm,
      inner sep=0pt, anchor=south west] (legloop) at (legA) {};
\node[font=\scriptsize, text=black!68, anchor=west]
    at ([xshift=0.10cm]legloop.east) {generated by loop (b)};

\node[detbox, fill=white, minimum width=0.34cm, minimum height=0.20cm,
      inner sep=0pt, anchor=south west, xshift=0.015cm] (legdet)
    at ([yshift=-0.30cm]legA) {};
\node[font=\scriptsize, text=black!68, anchor=west]
    at ([xshift=0.10cm]legdet.east) {generated deterministically};

% =========================
% Panel b
% =========================
\coordinate (bNW) at (11.35-1.3, 1.20);
\coordinate (bSE) at (16.85-1.3,-1.20);
\coordinate (bin) at ($(bNW)+(-0.42,-1.30)$);
\coordinate (bout) at ($(bSE)+(0.42,1.10)$);

\draw[loopframe] (bNW) rectangle (bSE);

\node[plain] (gen) at (12.75-1.3,-0.10) {Generator};
\node[plain] (ver) at (15.45-1.3,-0.10) {Verifier $\times K$};

\draw[flow] (bin) -- (gen.west);
\draw[flow] (gen.east) -- (ver.west);
\draw[flow] (ver.east) -- (bout);

\coordinate (genTop) at ($(gen.north)+(0,0.40)$);
\coordinate (verTop) at ($(ver.north)+(0,0.40)$);
\draw[feedbackflow]
    (ver.north) -- (verTop)
    -- node[above, feedbacklabel, yshift=-0.5mm] {feedback} (genTop)
    -- (gen.north);

\node[paneltag, anchor=south west]
    at ([xshift=-1mm,yshift=-0.3mm]bNW) {b)};

\end{tikzpicture}%
}
\caption{\textbf{Problem generation pipeline.} \textbf{a)} Each problem is seeded with a unique incident report. The report is converted into an incident model, which is then used to generate both a reference answer and an incident simulator; running the simulator produces the log file. The question template is fixed and shared across problems. Together, the question, logs, and reference answer constitute one benchmark problem. Solid outlines mark objects produced by the generator--verifier loop; double outlines mark deterministic outputs. \textbf{b)} The generator--verifier loop alternates between proposal and verification: the generator proposes an object, the verifier checks it, and any failed check returns feedback for revision. An object is accepted only after passing verification in $K$ consecutive rounds.}
\label{fig:slsbench_generation}
\end{figure*}
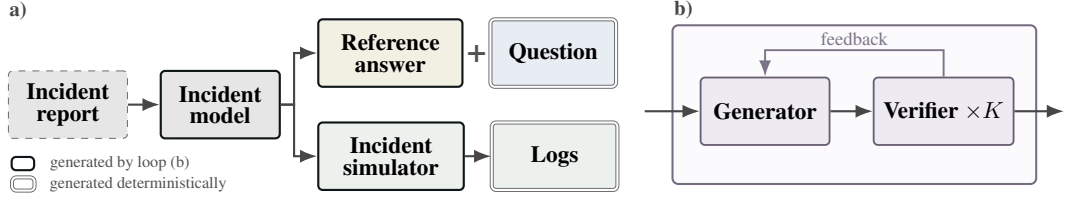

\paragraph{Incident model.}
While incident reports describe the overall shape and progression of an incident, they frequently omit critical low-level details necessary for log and summary generation, such as comprehensive component lists, inter-component interactions, and detailed log emission behaviors during the incident. We infer these missing details from the provided report context.

We introduce the \emph{incident model} (Fig.~\ref{fig:incident_model_schematic}) as a structured framework for interpreting the incident reports and systematically inferring the missing information from them in a consistent, controllable, and verifiable manner. The incident model is the core of our pipeline and drives the generation of both incident logs and summaries (Fig.~\ref{fig:slsbench_generation}a). It comprises three main parts that we introduce next.

First, the \emph{system structure} (Fig.~\ref{fig:incident_model_schematic}a) defines a static graph of log-emitting components, specifying the system's topology and per-component behavior. For each component, we define its name, interactions with other components, and the log templates it emits. A log template specifies the structure of a log line using fixed and variable slots (e.g., \texttt{user * logged in}). Each variable takes its allowed values, such as an integer range, a predefined set, or a regular expression (e.g., an IP address). Emitting a log line involves instantiating these variables with concrete values (e.g., \texttt{user Alice logged in}) and appending the result to the log stream. Components can represent individual processes, services, or sub-systems, and can be assigned to multiple hosts to simulate distributed environments.

Second, \emph{log emission mechanisms} (Fig.~\ref{fig:incident_model_schematic}b) govern which templates are emitted and when:
\begin{enumerate}
    \item 
\emph{Burst emissions} generate a specified number of logs from a single template within a short interval, modeling events such as configuration changes, restarts, or transient errors.
    \item 
\emph{Background emissions} generate logs from a single template at a fixed rate, modeling recurring activity such as health checks or baseline traffic.
    \item 
\emph{Path emissions} generate an ordered sequence of logs across multiple templates at a fixed rate with per-hop latencies, used to simulate requests traversing multiple system components.
\end{enumerate}
We selected these mechanisms based on template-emission patterns observed in real log streams. App.~\ref{app:incident-model-format} specifies their parameters and details their definitions within the incident model.

Third, the \emph{phased timeline} (Fig.~\ref{fig:incident_model_schematic}c) defines the temporal structure of the incident, partitioned into three phase types: a \emph{normal} phase establishing the baseline behavior, a \emph{trigger} phase introducing the failure-inducing event, and one or more \emph{progression} phases detailing the failure's evolution. Each phase specifies active emission mechanisms and their respective intensities, allowing log templates to emerge, vanish, or change emission rate as the incident progresses. For path emissions, phases also modulate $p50$ and $p95$ inter-hop latencies. Emission properties remain constant within a given phase, creating stepwise shifts in log patterns when transitioning between phases.
% Third, the \emph{phased timeline} (Fig.~\ref{fig:incident_model_schematic}c) defines the temporal structure of the incident, partitioned into three phase types: a \emph{normal} phase establishing baseline behavior, a \emph{trigger} phase introducing the failure-inducing event, and one or more \emph{progression} phases detailing the failure's evolution. Each phase dictates which emission mechanisms are active and their respective intensities, allowing log templates to emerge, vanish, or fluctuate in frequency as the incident progresses. For path emissions, phases also modulate $p50$ and $p95$ inter-hop latencies, enabling simulated performance degradation. Emission properties remain constant within a given phase, creating stepwise shifts in log patterns when transitioning between phases.

In summary, the incident model formally specifies the system topology, available log templates, and temporal evolution of log emission patterns. This framework allows us to translate diverse, often underspecified real-world incident reports into a unified, well-specified representation. Missing information is inferred in a manner consistent with the overall model, enabling controlled data generation, systematic evaluation, and clear interfaces for downstream pipeline stages.

We leverage an LLM to translate raw incident reports into these structured incident models. The LLM is instructed to preserve the core dynamics and structure of the original incident while coherently inferring missing elements, such as adding plausible system components and defining appropriate log templates for specific technologies. Rather than attempting to reconstruct the entire production architecture, we focus on inferring the minimal system subset necessary to contextualize the failure. The quality of these generated models is subsequently validated using a combination of algorithmic checks and LLM-based verifiers (Sec.~\ref{sec:generation}). The final incident models are serialized as YAML files (see App.~\ref{app:incident-model-format} for an example), capturing all three components discussed above. Aggregate statistics characterizing these generated models are provided in App.~\ref{app:incident-model-stats}.

\input{figures/3-incident-model-figure}

\paragraph{Incident simulator and log generation.}
\label{sec:simulator}

While the incident model specifies the system topology and phase-dependent emission mechanisms, it does not yield a concrete log stream. Specifically, it abstracts away exact timestamps, emitted log identities, realized template variables, and the granular execution of encoded behaviors. To instantiate a concrete log stream, the incident model must be executed by a simulator that operationalizes these high-level specifications into low-level details. For each incident model, we leverage an LLM to generate a deterministic Python-based simulator that adheres to the model specification and coordinates the following low-level aspects to produce logs:

\begin{itemize}
    \item \textbf{Timing and counts.} The simulator translates emission mechanisms into discrete events with precise timestamps. For each phase, it determines the scheduling of bursts, background, and path emissions, and enforces specified inter-log latencies along request paths.
    \item \textbf{Variable values.} Upon emission, each template variable is populated with a value from its defined domain. These values are dynamically restricted to be contextually appropriate for the current incident phase (e.g., HTTP status codes remaining in the \texttt{2xx} range during normal operation, but shifting to \texttt{4xx} and \texttt{5xx} during a failure).
    \item \textbf{Variable coherence.} The simulator ensures that variable values remain logically valid and mutually consistent throughout a path execution. For instance, a log message indicating a connection timeout must not co-occur with a variable indicating a fast connection time; similarly, if a request path registers a failure, subsequent logs within that same path cannot report a successful completion.
\end{itemize}

The resulting simulator deterministically outputs a CSV file containing the log stream, typically ranging from 20k to 100k logs (56k on average). Artifact quality is enforced via algorithmic and LLM verifiers (Sec.~\ref{sec:generation}). A complete simulator example is provided in App.~\ref{app:simulator-example}. 

\paragraph{Question and reference answer.}
% Every problem in \slsbench\ uses the standardized prompt described in Section~\ref{sec:task-definition} and shown in Appendix~\ref{app:fixed-question-and-example-answer}. The reference answer follows the same format as the requested agent answer. Rather than generating reference summaries directly from the raw synthetic logs, we derive them from the incident model itself. Because the model compactly encapsulates the information needed for the task, the generator translates this structured representation into a precise narrative that satisfies the question prompt.

% Deriving summaries from the incident model risks incorporating ``invisible'' ground truth---internal model mechanics (e.g., a silent hardware fault) that never surface in the log stream. To prevent this, the generator is strictly instructed to describe only the log-observable implications of internal processes, a constraint supported by the incident model’s explicit mapping of behaviors to log emissions. The resulting 90 reference summaries range from 1,900 to 3,300 words (2,400 on average). Reference answer quality is assessed using an LLM verifier (Sec.~\ref{sec:generation}).

Every problem in \slsbench\ uses a standardized question requiring an agent to analyze the logs and produce an incident summary with four components: (i) incident highlight bullets; (ii) system structure and behavior; (iii) phase-by-phase walkthrough detailing observed behaviors, interpretations, evidence, and confidence; and (iv) an overarching diagnosis specifying the root cause, causal chain, evidence, and confidence level (see App.~\ref{app:fixed-question-and-example-answer}).

We generate reference summaries directly from the incident model. Because the model compactly encapsulates all information about the incident, we instruct an LLM to translate this structured representation into a precise narrative that answers the provided question to produce the reference summary. However, deriving summaries from the incident model risks incorporating 	``invisible'' ground truth---internal model mechanics (e.g., a silent hardware fault) that never surface in the log stream. To prevent this, the generator is strictly instructed to describe only the log-observable implications of internal processes, a constraint natively supported by the incident model’s explicit mapping of behaviors to log emissions. The resulting 90 reference summaries range from 1,900 to 3,300 words (2,400 on average). Their quality is assessed using an LLM verifier (Sec.~\ref{sec:generation}).

\subsection{Pipeline Component Generation}
\label{sec:generation}

This section describes the generation pipeline for incident models, simulators, and reference answers introduced in Sec.~\ref{sec:components}, and covers the design choices that promote data quality and log realism. The code and prompts used in the pipeline are provided in our code repository.

\paragraph{Generation approach.}
We generate all benchmark artifacts using the generator--verifier loop in Fig.~\ref{fig:slsbench_generation}b. First, a generator proposes an object, which a verifier evaluates against object-specific criteria. If no violations are detected, the object is accepted; otherwise, the verifier provides constructive feedback, prompting the generator to revise the object. We allow up to 20 such iterations per object. If the generator fails to produce a valid artifact within this budget, the process restarts from scratch.

Verification depends on the artifact type. Incident models and simulators undergo both algorithmic and semantic validation. Algorithmic verifiers enforce output schema adherence, structural validity, and internal consistency. For incident models, LLM verifiers receive the source report and check faithfulness to its stated facts and the plausibility of inferred details (App.~\ref{app:incident-model-validity}); simulator verifiers check adherence to the incident model.  Reference answers, whose requirements are predominantly semantic (e.g., narrative faithfulness to the incident model), are evaluated solely by an LLM verifier.

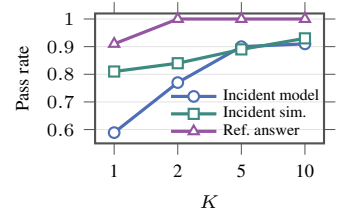
\begin{wrapfigure}[10]{r}{0.35\textwidth}
    \centering
    \vspace{-10pt}
    % --- color definitions (muted Dynatrace palette) ---
% Place these in your preamble or near the top of the figure.
\definecolor{palblue}{HTML}{4E6FB5}     % muted primary
\definecolor{palpurple}{HTML}{6E5FA8}   % muted secondary (unused here, kept for reference)
\definecolor{palmagenta}{HTML}{9655A0}  % muted accent
\definecolor{palteal}{HTML}{3D8B7E}     % muted teal

\centering
\begin{tikzpicture}
\begin{axis}[
    width=0.33\textwidth,
    height=3.3cm,
    symbolic x coords={1,2,5,10},
    xtick=data,
    ymin=0.55,
    ymax=1.02,
    ytick={0.6,0.7,0.8,0.9,1.0},
    axis lines=box,
    xlabel={$K$},
    ylabel={Pass rate},
    xmajorgrids=false,
    ymajorgrids=true,
    grid style={draw=outergray!40, line width=0.35pt},
    tick align=outside,
    axis line style={draw=bordergray, line width=0.8pt},
    tick style={draw=bordergray},
    tick label style={font=\scriptsize},
    label style={font=\scriptsize},
    legend style={
        draw=none,
        fill=none,
        font=\tiny,
        at={(1.04,-0.05)},
        anchor=south east,
        cells={anchor=west},
        row sep=-3.0pt
    },
    clip=false,
]
\addplot+[
    draw=palblue,
    line width=1.2pt,
    mark=*,
    mark size=2.0pt,
    mark options={draw=palblue, fill=white, line width=0.8pt}
] coordinates {
    (1,0.5889)
    (2,0.7700)
    (5,0.9000)
    (10,0.9100)
};
\addlegendentry{Incident model}
\addplot+[
    draw=palteal,
    line width=1.2pt,
    mark=square*,
    mark size=1.9pt,
    mark options={draw=palteal, fill=white, line width=0.8pt}
] coordinates {
    (1,0.8100)
    (2,0.8400)
    (5,0.8900)
    (10,0.9300)
};
\addlegendentry{Incident sim.}
\addplot+[
    draw=palmagenta,
    line width=1.2pt,
    mark=triangle*,
    mark size=2.2pt,
    mark options={draw=palmagenta, fill=white, line width=0.8pt}
] coordinates {
    (1,0.9100)
    (2,1.0000)
    (5,1.0000)
    (10,1.0000)
};
\addlegendentry{Ref.\ answer}
\end{axis}
\end{tikzpicture}
    \vspace{-5pt}
    \caption{Increasing the verifier threshold $K$ improves the pass rate.}
    \label{fig:verifier_k_sweep}
\end{wrapfigure}

Producing high-fidelity objects requires multiple independent verifier passes. As Fig.~\ref{fig:verifier_k_sweep} demonstrates, objects accepted after only a single pass ($K = 1$) frequently fail subsequent verifications (i.e., have low pass rate). Manual inspection reveals that $K = 1$ failures typically stem from genuine requirement violations that degrade quality. Conversely, failures at higher thresholds (e.g., $K = 10$) are often artifacts of overly rigid rubric interpretations that do not meaningfully impact object utility. Thus, consecutive verifications yield verifier-stable objects that, once accepted, are highly likely to pass again.

We employ GPT-5.2 (high) for both generation and verification. Strong verifiers are crucial for detecting subtle constraint violations; weaker generators often struggle to incorporate feedback from a stronger verifier within the iteration budget, resulting in long or failed object generation.

To evaluate the sensitivity of the LLM verifiers, we manually injected non-trivial requirement violations into 50 accepted objects ($K = 10$) for each artifact type. The verifiers successfully detected $98\%$ of these injected errors, confirming their high sensitivity to constraint violations.

\begin{wrapfigure}[15]{r}{0.40\linewidth}
  \centering
  \vspace{-10pt}
  \includegraphics[width=\linewidth]{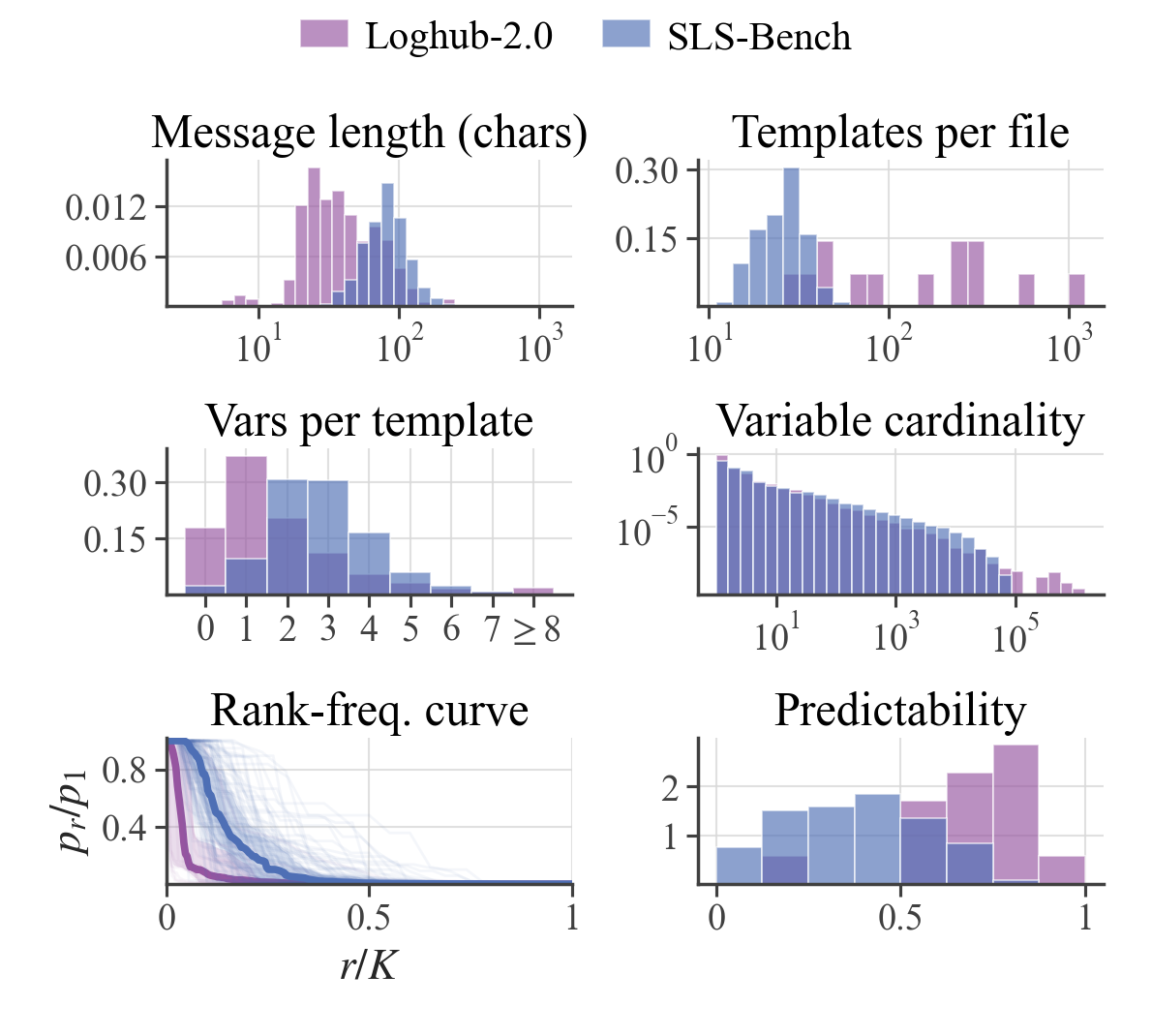}
  \vspace{-20pt}
  \caption{Comparing Loghub-2.0 and \slsbench\ on representative statistics.}
  \label{fig:log-stats-comparison}
\end{wrapfigure}

\paragraph{Generated log realism.}
While high-level realism is anchored by the source incident reports, low-level realism relies on the LLM generating contextually appropriate log templates. To assess this, we compare \slsbench\ against Loghub-2.0 (real logs from 14 systems) across metrics capturing message form, template diversity, and local stream structure (Fig.~\ref{fig:log-stats-comparison}; see App.~\ref{app:log-stat-comparison} for definitions).

\slsbench\ matches Loghub-2.0 in message length, variables per template, and variable cardinality, indicating that our individual log lines exhibit realistic structure and variable diversity. However, three notable differences emerge:
\begin{enumerate}
    \item \slsbench\ contains fewer unique templates per file. This is an intentional consequence of simulating the minimal sub-system relevant to the incident, rather than an entire production environment laden with unrelated background noise.
    \item While both datasets display heavy-tailed rank–frequency curves, \slsbench’s curves are flatter, indicating a more even distribution of template occurrences rather than extreme concentration in a few dominant templates.
    \item Adjacent-template predictability is lower in \slsbench. Because our synthetic streams interleave diverse background activities, concurrent request paths, and phase-specific failure bursts, they avoid the highly repetitive, predictable template cycles characteristic of many Loghub-2.0 datasets.
\end{enumerate}
Overall, \slsbench\ achieves message-level realism, with minor deviations in template diversity and sequence predictability directly attributable to our incident-centric generation design. However, high-level properties such as ambiguity, incomplete evidence, and conflicting signals in the log stream are harder to formalize and quantify; realism along these axes relies on grounding in real-world incident reports and adherence to internally consistent incident models (App.~\ref{app:incident-model-validity}).

\subsection{LLM Judge Scoring}
\label{sec:scoring}

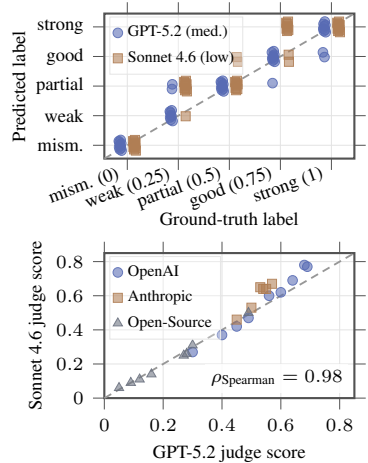
\begin{wrapfigure}[19]{r}{0.41\textwidth}
    \centering
    \vspace{-20pt}
    % Preamble additions (once):
%   \definecolor{othergray}{RGB}{107,114,128}

\begin{tikzpicture}
\begin{groupplot}[
    group style={
        group size=1 by 2,
        vertical sep=1.25cm,
    },
    width=0.35\textwidth,
    height=3.5cm,
    axis lines=box,
    xmajorgrids=true,
    ymajorgrids=true,
    grid style={draw=outergray!40, line width=0.35pt},
    tick align=outside,
    axis line style={draw=bordergray, line width=0.8pt},
    tick style={draw=bordergray},
    tick label style={font=\scriptsize},
    label style={font=\scriptsize},
    xlabel style={yshift=0.5mm},
    ylabel near ticks,
    ylabel style={yshift=-1mm, font=\scriptsize},
    legend style={
        draw=black!10,
        fill=white,
        fill opacity=0.3,
        text opacity=1,
        font=\tiny,
        at={(0.02,0.98)},
        anchor=north west,
        cells={anchor=west},
        inner sep=1.2pt,
        row sep=0pt
    },
]

% ═══════════════════════════════════════════════════════════════════════
% Panel (a) — per-label predictions
% ═══════════════════════════════════════════════════════════════════════
\nextgroupplot[
    xmin=0.55, xmax=5.45,
    ymin=0.55, ymax=5.45,
    xtick={1,2,3,4,5},
    xticklabels={mism. (0), weak (0.25), partial (0.5), good (0.75), strong (1)},
    ytick={1,2,3,4,5},
    yticklabels={mism., weak, partial, good, strong},
    xlabel={Ground-truth label},
    xlabel style={yshift=2mm},
    ylabel={Predicted label},
    xticklabel style={rotate=20, anchor=east},
    yticklabel style={rotate=0, anchor=east},
]

\addplot+[
    no markers, densely dashed, draw=bordergray!60, line width=0.7pt, forget plot
] coordinates {
    (0.55,0.55)
    (5.45,5.45)
};

\addplot+[
    only marks, mark=*, mark size=1.8pt, line width=0.3pt,
    mark options={draw=openaiblue, fill=openaiblue, fill opacity=0.58, draw opacity=0.92}
] coordinates {
    (4.834,4.14) (4.84,4.94) (4.846,5.1)  (4.852,4.86) (4.858,5.02)
    (4.864,3.98) (4.87,5.18) (4.876,4.9)  (4.882,5.06) (4.888,4.82)
    (3.834,4.14) (3.84,3.94) (3.846,3.1)  (3.852,3.86) (3.858,4.02)
    (3.864,3.98) (3.87,4.18) (3.876,3.9)  (3.882,4.06) (3.888,3.82)
    (2.834,3.14) (2.84,2.94) (2.846,3.1)  (2.852,2.86) (2.858,3.02)
    (2.864,2.98) (2.87,3.18) (2.876,2.9)  (2.882,3.06) (2.888,2.82)
    (1.834,2.14) (1.84,1.94) (1.846,2.1)  (1.852,1.86) (1.858,2.02)
    (1.864,1.98) (1.87,2.18) (1.876,2.9)  (1.882,3.06) (1.888,1.82)
    (0.834,1.14) (0.84,0.94) (0.846,1.1)  (0.852,0.86) (0.858,1.02)
    (0.864,0.98) (0.87,1.18) (0.876,0.9)  (0.882,1.06) (0.888,0.82)
};
\addlegendentry{GPT-5.2 (med.)}

\addplot+[
    only marks, mark=square*, mark size=1.7pt, line width=0.3pt,
    mark options={draw=anthropicorange, fill=anthropicorange, fill opacity=0.58, draw opacity=0.92}
] coordinates {
    (5.114,5.14) (5.12,4.94) (5.126,5.1) (5.132,4.86) (5.138,5.02)
    (5.144,4.98) (5.15,5.18)  (5.156,4.9) (5.162,5.06) (5.168,4.82)
    (4.114,5.14) (4.12,4.94) (4.126,5.1) (4.132,4.86) (4.138,5.02)
    (4.144,4.98) (4.15,5.18)  (4.156,4.9) (4.162,4.06) (4.168,3.82)
    (3.114,3.14) (3.12,2.94) (3.126,3.1) (3.132,2.86) (3.138,3.02)
    (3.144,3.98) (3.15,3.18)  (3.156,2.9) (3.162,3.06) (3.168,3.82)
    (2.114,3.14) (2.12,2.94) (2.126,3.1) (2.132,2.86) (2.138,3.02)
    (2.144,1.98) (2.15,3.18)  (2.156,2.9) (2.162,3.06) (2.168,2.82)
    (1.114,1.14) (1.12,0.94) (1.126,1.1) (1.132,0.86) (1.138,1.02)
    (1.144,0.98) (1.15,1.18)  (1.156,0.9) (1.162,1.06) (1.168,0.82)
};
\addlegendentry{Sonnet 4.6 (low)}

% ═══════════════════════════════════════════════════════════════════════
% Panel (b) — GPT vs Claude judge agreement
% ═══════════════════════════════════════════════════════════════════════
\nextgroupplot[
    xmin=0, xmax=0.85,
    ymin=0, ymax=0.85,
    xtick={0,0.2,0.4,0.6,0.8},
    ytick={0,0.2,0.4,0.6,0.8},
    xlabel={GPT-5.2 judge score},
    ylabel={Sonnet 4.6 judge score},
]

\addplot+[
    no markers, densely dashed, draw=bordergray!60, line width=0.7pt, forget plot
] coordinates {
    (0,0)
    (0.85,0.85)
};

\addplot+[
    only marks, mark=*, mark size=1.8pt, line width=0.3pt,
    mark options={draw=openaiblue, fill=openaiblue, fill opacity=0.58, draw opacity=0.92}
] coordinates {
    (0.68,0.78) (0.64,0.69) (0.69,0.77)
    (0.60,0.62) (0.49,0.47) (0.56,0.60)
    (0.45,0.42) (0.30,0.27) (0.40,0.37)
};
\addlegendentry{OpenAI}

\addplot+[
    only marks, mark=square*, mark size=1.7pt, line width=0.3pt,
    mark options={draw=anthropicorange, fill=anthropicorange, fill opacity=0.58, draw opacity=0.92}
] coordinates {
    (0.45,0.46) (0.57,0.67) (0.50,0.53)
    (0.55,0.64) (0.53,0.65) (0.54,0.64)
};
\addlegendentry{Anthropic}

\addplot+[
    only marks, mark=triangle*, mark size=2.0pt, line width=0.3pt,
    mark options={draw=othergray, fill=othergray, fill opacity=0.58, draw opacity=0.92}
] coordinates {
    (0.27,0.25) (0.49,0.50) (0.30,0.31)
    (0.12,0.11) (0.16,0.14) (0.05,0.06)
    (0.28,0.26) (0.09,0.09)
};
\addlegendentry{Open-Source}

\node[
    anchor=south east,
    font=\scriptsize,
    fill=white,
    fill opacity=0.85,
    text opacity=1,
    inner sep=2pt
] at (axis cs:0.83,0.03) {$\rho_{\text{Spearman}} = 0.98$};

\end{groupplot}
\end{tikzpicture}
    \vspace{-5pt}
    \caption{\textbf{Top:} Judge prediction vs. ground-truth on calibration set. \textbf{Bottom:} Judge agreement across evaluated LLMs.}
    \label{fig:judge_label_scatter}
\end{wrapfigure}
To evaluate agent performance, we employ an LLM-as-a-judge paradigm. The judge receives the standardized question, the reference summary, and the agent's predicted summary, and assigns one of five labels: \emph{mismatch}, \emph{weak match}, \emph{partial match}, \emph{good match}, or \emph{strong match}. The grading rubric spans seven dimensions evaluating both structural compliance and content accuracy, and produces the final score by combining the per-dimension scores (see App.~\ref{app:judge-rubric} for full definitions).

To select a judge, we constructed a calibration dataset utilizing 10 benchmark problems. For each problem, we generated five candidate predictions corresponding to the five rubric labels. The \emph{strong match} was generated as an LLM paraphrase of the reference answer. \emph{Good}, \emph{partial}, and \emph{weak} matches were derived by progressively degrading this paraphrase to intentionally violate specific rubric thresholds. The \emph{mismatch} baseline was created by combining a problem with a reference answer from an entirely different incident.

Based on the achieved calibration accuracy of $90\%$, GPT-5.2 (medium) was selected as the primary judge for the evaluation. However, large-scale benchmark runs revealed a slight strictness bias, where GPT-5.2 occasionally penalized harmless additions, such as agents reporting extra log-value details. Refining the prompt rubric did not mitigate this behavior. To counterbalance this strictness and reduce model-family bias, we introduced a second judge: Sonnet 4.6 (low). Sonnet demonstrated greater leniency toward harmless additions, though it was comparatively more prone to missing genuine rubric violations, resulting in a lower independent calibration accuracy ($62\%$).

To compute the final score, we map the five rubric labels to continuous values ($0$, $0.25$, $0.50$, $0.75$, and $1.00$) and average the predictions of both judges. Fig.~\ref{fig:judge_label_scatter} (top) illustrates per-label calibration performance: both judges reliably identify \emph{mismatch}, \emph{partial match}, and \emph{strong match} categories, while GPT-5.2 excels at distinguishing the boundary between \emph{weak match} and \emph{good match}. Fig.~\ref{fig:judge_label_scatter} (bottom) shows that across 14 evaluated LLMs, the mean scores of the two judges differ by no more than $0.12$, and their final model rankings remain highly correlated ($\rho_{\text{Spearman}} = 0.98$). We further test judge robustness, agreement with human evaluation, and external log analytics benchmarks in App.~\ref{app:benchmark-validity}.

%\newpage
\section{Results}
\label{sec:results}

To establish a baseline on \slsbench,\ we evaluate an LLM agent equipped with a stateful Python interpreter for analyzing logs using \texttt{pandas} \citep{mckinney2010data} and standard Python packages. Our evaluation includes closed-source frontier models, specifically the GPT-5.4 \citep{openai2026gpt54} and Claude 4.x models \citep{anthropic2026opus46}, and a diverse set of open-weight models, including Ministral 3 14B and Mistral Large 3 \citep{mistral2025mistral3}, Nemotron 3 Nano and Nemotron 3 Super \citep{nvidia2025nemotron3}, Qwen3-Next-80B-A3B \citep{qwen2025qwen3}, MiniMax-M2 \citep{minimax2025m2}, GLM-5 \citep{glm2026glm5}, and DeepSeek-V3.2 \citep{deepseek2025v32}. All prompts, the agent implementation, and evaluation scripts are open-sourced in our repository, with further details provided in App.~\ref{app:eval_details}.

\begin{table}[t]
\centering
\caption{\textbf{Benchmark scores with cost and time per problem.} Score is mean $\pm$ SEM, time is mean $\pm$ SD; cost and time are averages per problem. Open-weight model parameters are total/active for mixture-of-expert models. A $\star$\, denotes Pareto-optimal models under the cost--score trade-off (Fig.~\ref{fig:results_all_models}). We use the sample variance formula to compute standard deviation and standard error of the mean.}
\label{tab:model_cost_perf}
\footnotesize
\renewcommand{\arraystretch}{0.92}
\setlength{\tabcolsep}{4pt}
\begin{tabular}{@{}l l S[table-format=2.2] c S[table-format=2.2] S[table-format=1.2]@{}}
\toprule
Model
  & \multicolumn{1}{c}{Score ($\uparrow$)}
  & \multicolumn{1}{c}{Cost [\$] ($\downarrow$)}
  & \multicolumn{1}{c}{Time [Min] ($\downarrow$)}
  & \multicolumn{1}{c}{Score/\$}
  & \multicolumn{1}{c}{Score/Min} \\
\midrule
\multicolumn{6}{@{}l}{\textit{OpenAI}} \\
GPT-5.4 (high)                  & $0.73 \pm 0.01$ & 2.19 & $9.5 \pm 1.6$  &  0.33 & 0.08 \\
GPT-5.4 (medium)$^{\star}$      & $0.73 \pm 0.01$ & 1.56 & $6.9 \pm 1.0$  &  0.47 & 0.10 \\
GPT-5.4 (low)$^{\star}$         & $0.67 \pm 0.01$ & 0.85 & $3.4 \pm 0.5$  &  0.79 & 0.19 \\
GPT-5.4 Mini (high)$^{\star}$   & $0.61 \pm 0.01$ & 0.37 & $5.2 \pm 1.3$  &  1.65 & 0.11 \\
GPT-5.4 Mini (medium)$^{\star}$ & $0.57 \pm 0.01$ & 0.21 & $3.0 \pm 0.7$  &  2.71 & 0.18 \\
GPT-5.4 Mini (low)$^{\star}$    & $0.48 \pm 0.01$ & 0.06 & $0.9 \pm 0.1$  &  8.00 & 0.53 \\
GPT-5.4 Nano (high)             & $0.44 \pm 0.01$ & 0.10 & $7.0 \pm 1.7$  &  4.40 & 0.06 \\
GPT-5.4 Nano (medium)$^{\star}$ & $0.39 \pm 0.01$ & 0.06 & $3.7 \pm 0.9$  &  6.50 & 0.09 \\
GPT-5.4 Nano (low)$^{\star}$    & $0.28 \pm 0.01$ & 0.01 & $1.1 \pm 0.2$  & 28.00 & 0.22 \\
\addlinespace[3pt]
\multicolumn{6}{@{}l}{\textit{Anthropic}} \\
Opus 4.6 (high)                 & $0.61 \pm 0.01$ & 1.91 & $9.9 \pm 2.9$  &  0.32 & 0.06 \\
Opus 4.6 (medium)               & $0.59 \pm 0.01$ & 0.71 & $4.1 \pm 0.8$  &  0.83 & 0.13 \\
Sonnet 4.6 (medium)             & $0.57 \pm 0.01$ & 1.02 & $11.1 \pm 3.5$ &  0.56 & 0.05 \\
Sonnet 4.6 (low)                & $0.58 \pm 0.01$ & 0.47 & $5.0 \pm 1.4$  &  1.23 & 0.11 \\
Opus 4.6 (low)                  & $0.53 \pm 0.01$ & 0.34 & $2.5 \pm 0.5$  &  1.56 & 0.21 \\
Haiku 4.5                       & $0.46 \pm 0.01$ & 0.16 & $2.4 \pm 0.5$  &  2.88 & 0.19 \\
\addlinespace[3pt]
\multicolumn{6}{@{}l}{\textit{Open--Weight}} \\
GLM-5 \tiny(744B/40B)                  & $0.49 \pm 0.01$ & 0.59 & $7.9 \pm 3.5$  &  0.83 & 0.05 \\
MiniMax-M2 \tiny(230B/10B)$^{\star}$   & $0.29 \pm 0.01$ & 0.02 & $3.4 \pm 1.1$  & 14.50 & 0.06 \\
Nemotron 3 Super \tiny(120B/12B)       & $0.29 \pm 0.02$ & 0.02 & $7.8 \pm 5.1$  & 14.50 & 0.03 \\
DeepSeek V3.2 \tiny(685B/37B)          & $0.27 \pm 0.02$ & 0.25 & $11.8 \pm 6.1$ &  1.08 & 0.02 \\
Mistral Large 3 \tiny(675B/41B)        & $0.12 \pm 0.01$ & 0.02 & $0.8 \pm 0.4$  &  6.00 & 0.13 \\
Ministral 3 \tiny(14B)                 & $0.12 \pm 0.01$ & 0.02 & $1.8 \pm 1.1$  &  6.00 & 0.05 \\
Qwen3-Next \tiny(80B/3B)               & $0.10 \pm 0.01$ & 0.02 & $3.7 \pm 1.9$  &  5.00 & 0.02 \\
Nemotron 3 Nano \tiny(30B/3.5B)        & $0.06 \pm 0.01$ & 0.02 & $5.1 \pm 5.1$  &  3.00 & 0.01 \\
\bottomrule
\end{tabular}
\end{table}

\paragraph{Main results.} Table~\ref{tab:model_cost_perf} presents evaluation results for all models. We report the mean score across the 90 benchmark problems, along with the average monetary cost and runtime per problem. OpenAI models define most of the Pareto frontier in the cost--performance trade-off space, with MiniMax-M2 also Pareto-optimal at low cost. Among closed-source models, performance generally scales with model size and reasoning effort. However, within any given size class, models using lower reasoning effort offer the best cost- and time-efficiency ratios. Open-weight models currently lag behind their closed-source counterparts, with GLM-5 achieving the strongest open-weight score. Our LLM judge rubric maps \emph{good match} to 0.75, which we use as the threshold for a reliable incident reconstruction. The strongest models approach this threshold, with GPT-5.4 medium and high both reaching 0.73 on average, requiring \$1.56--\$2.19 and 6.9--9.5 minutes per problem on average. App. \ref{app:tool-call-budget} shows model rankings are stable under different tool-call budgets and that the scores increase with the budget.

\begin{wrapfigure}[16]{r}{0.45\linewidth}
    \centering
    \includegraphics[width=1\linewidth]{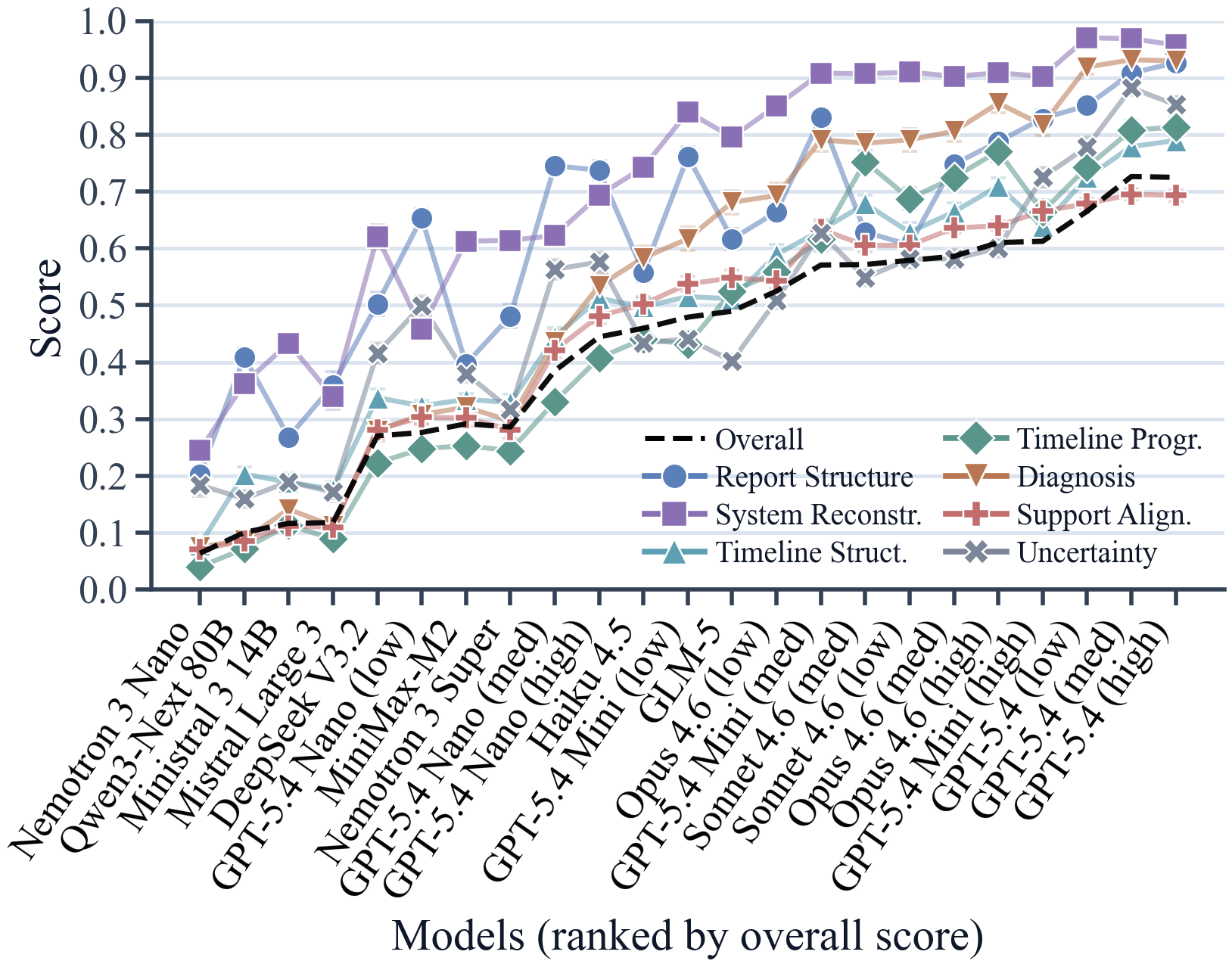}
    \caption{Overall score split into its constituent axes as evaluated by the LLM judge.}
    \label{fig:all_and_per_dim_results}
\end{wrapfigure}
\paragraph{Disaggregated score analysis.} 
Fig.~\ref{fig:all_and_per_dim_results} shows the disaggregation of the overall score into its seven constituent rubric dimensions. 
% (defined in Appendix~\ref{app:judge-rubric}).
Per-axis scores generally improve with the overall score, model capability, and reasoning effort. The score breakdown shows why the task remains difficult. \emph{Diagnosis}, which evaluates high-level incident chain reconstruction, is nearing saturation at the maximum score of 1. In contrast, \emph{timeline progression} and \emph{support alignment}, which assess whether high-level reconstructions are supported by granular log evidence, remain unsaturated and typically plateau in the 0.7--0.8 range. Error analysis of GPT-5.4 (medium) predictions with the lowest support alignment scores shows that the main failure modes involve omitting or misinterpreting the critical evidence used by the reference answer.

\begin{wrapfigure}[19]{r}{0.35\linewidth}
    \centering
    \vspace{-1em}
    \begin{subfigure}{\linewidth}
        \centering
        \includegraphics[width=\linewidth]{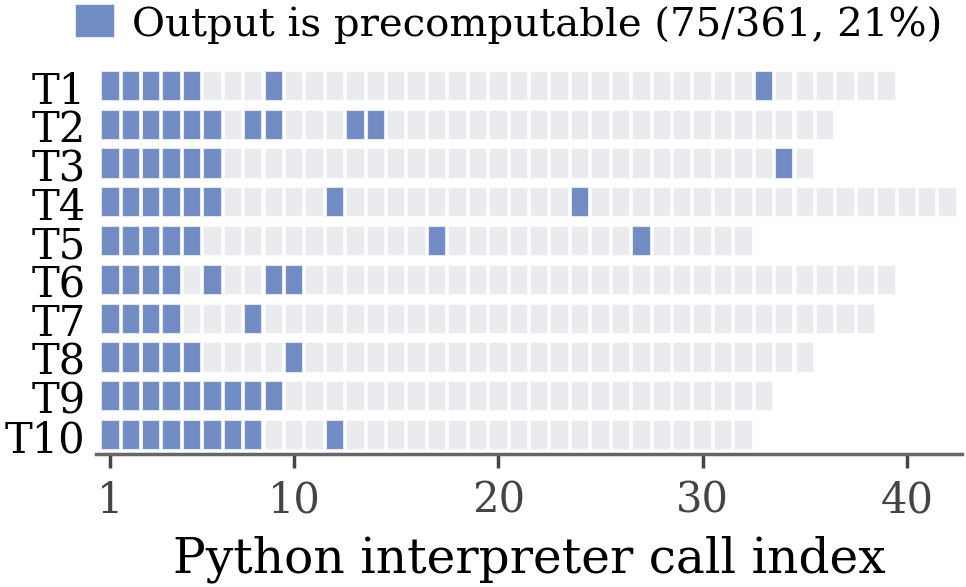}
        \label{fig:precompute_opportunity}
    \end{subfigure}
    
    \vspace{0.0em}
    
    \begin{subfigure}{\linewidth}
        \centering
        \includegraphics[width=\linewidth]{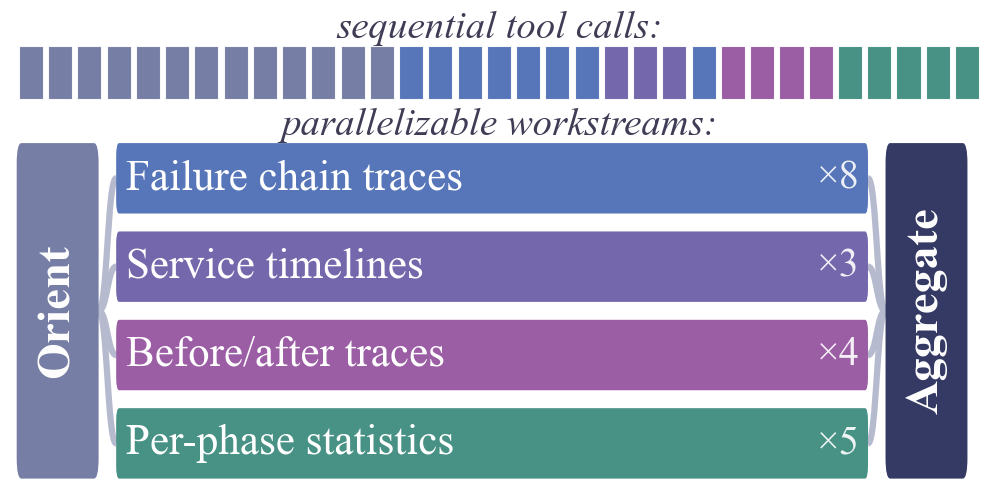}
        \label{fig:parallel_workstreams}
    \end{subfigure}
    \vspace{-1.9em}
    \caption{
    % \textbf{Efficiency improvement opportunities.}
    \textbf{Top:} Precomputable Python tool outputs. \textbf{Bottom:} After initial orientation, sequential tool calls can be parallelized into independent streams.}
    \label{fig:efficiency_analysis}
\end{wrapfigure}
\paragraph{Efficiency analysis.} 
% \slsbench\ also supports analyses beyond model ranking.
Top-scoring LLMs can generate useful incident summaries, but they do so with significant cost and latency overheads. To identify optimization opportunities, we analyzed 10 reasoning trajectories from GPT-5.4 (medium). These trajectories consistently begin with a generic orientation phase (Fig.~\ref{fig:efficiency_analysis}, top), during which the agent uses the Python interpreter to inspect the log schema, size, time span, unique hosts and services, errors, and templates, while also constructing temporal views (e.g., per-minute volume, severity trends) to localize the incident. Across the analyzed trajectories, 21\% (75 out of 361) of the Python tool outputs were precomputable, suggesting that this context could be injected prior to the agent's analysis to reduce API calls. Furthermore, measured by tool-output count, the initial orientation and planning prefix accounts for roughly 30\% of an average trajectory. After this prefix, the agent typically branches into independent analytical streams that could be effectively parallelized (Fig.~\ref{fig:efficiency_analysis}, bottom). Further details are available in App.~\ref{app:efficiency-analysis}.

% \begin{figure}[t]
%     \centering
%     \begin{subfigure}{\linewidth}
%         \centering
%         \includegraphics[width=0.33\linewidth]{figures/precompute_opportunity.png}
%         \label{fig:precompute_opportunity}
%     \end{subfigure}
%     \begin{subfigure}{\linewidth}
%         \centering
%         \includegraphics[width=0.33\linewidth]{figures/parallel_workstreams.png}
%         \label{fig:parallel_workstreams}
%     \end{subfigure}
%     \caption{\textbf{Efficiency opportunities within GPT-5.4 (medium) reasoning trajectories.} (Top) Python tool-output events containing generic log artifacts that can be precomputed before the agent begins its reasoning process. (Bottom) Following an initial orientation and planning prefix, the sequential tool-call chain can be parallelized into independent workstreams, with results aggregated prior to the final report synthesis.}
%     \label{fig:efficiency_analysis}
% \end{figure}

\section{Conclusions}
\label{sec:conclusion}
In this work, we introduced \slsbench, a benchmark for evaluating LLM agents on incident log summarization from large-scale log streams. To construct \slsbench, we developed a synthetic data generation pipeline that converts incident reports into synthetic log streams and reference incident summaries. This pipeline demonstrates how real-world grounding, explicit incident modeling, strict validation, and statistical comparisons with real log streams can be used to produce realistic and controllable observability data. The resulting benchmark contains 90 distinct tasks, each derived from a unique real-world incident report. Our evaluation of 14 LLMs shows that current frontier models can produce viable incident summaries in this setting, but remain costly and slow, while open-weight models lag substantially behind in summary quality.

% \slsbench\ currently has two main limitations. First, coverage is limited to 90 publicly available incident reports; future versions can expand this pool with new public, private, or synthetic incident reports. Second, the incident model caps the complexity of the behavior each benchmark problem represents. While we include basic modeling components to represent realistic system behaviors, future work can extend the model to target richer systems, additional failure modes, or specific behaviors, such as background noise and the incidental complexity of complete production environments.

\slsbench\ currently has three main limitations. First, coverage is limited to 90 publicly available incident reports; future versions can expand this pool with new public, private, or synthetic incident reports. Second, the incident model caps the complexity of the behavior each benchmark problem represents. While we include basic modeling components to represent realistic system behaviors, future work can extend the model to target richer systems, additional failure modes, or specific behaviors, such as background noise and the incidental complexity of complete production environments. Third, logs and reference summaries are generated from the same incident model, and judges score predictions against the references without accessing the logs. Evaluation therefore depends on the references faithfully representing the evidence available in the generated logs. Judge-robustness tests and a small human study support this scoring approach, and rank agreement with o11y-bench and OpenRCA provides evidence of relevance to related observability tasks (App.~\ref{app:benchmark-validity}). Evaluating direct transfer of \slsbench\ scores to real incident log summarization is a natural next step in extending this validation.

Ultimately, \slsbench\ enables the systematic evaluation and improvement of LLM agents on realistic incident summarization tasks, while our pipeline offers a systematic and scalable blueprint for building future synthetic observability datasets. To support reproducible generation, evaluation, and future extensions, we have publicly released our data and code.

\bibliographystyle{plainnat}   % or: abbrvnat, unsrtnat, apalike, etc.
\bibliography{refs}            % points to refs.bib

%%%%%%%%%%%%%%%%%%%%%%%%%%%%%%%%%%%%%%%%%%%%%%%%%%%%%%%%%%%%

\clearpage
\appendix

\section*{Appendix Overview}
The following appendices provide supplementary technical details to support the main paper. Appendix~\ref{app:filtering-prompt} describes incident report collection, filtering, and licensing, and Appendix~\ref{app:incident-model-and-simulator-details} details the incident model and the simulator. Appendix~\ref{app:log-stat-comparison} provides additional statistical analyses. Appendix~\ref{app:fixed-question-and-example-answer} presents the fixed question prompt and an example reference answer, while Appendix~\ref{app:judge-rubric} provides a granular breakdown of our scoring methodology and label mappings. Appendix~\ref{app:dataset_constr_cost} reports token usage and costs across the generator and verifier steps of the data generation pipeline. Appendix~\ref{app:eval_details} describes our agent setup and inference configuration. Appendix~\ref{app:tool-call-budget} reports tool-call-budget sensitivity, and Appendix~\ref{app:efficiency-analysis} analyzes agent trajectories to identify cost and latency optimization opportunities. Appendix~\ref{app:benchmark-validity} reports judge robustness, human evaluation, and comparisons with external benchmarks.

\section{Incident Report Collection, Filtering, and Licensing}
\label{app:filtering-prompt}

\subsection{Collection and preprocessing}
\label{app:report-collection}

We collected candidate URLs through manual and LLM-assisted web search using broad queries such as ``incident report'', ``postmortem'', ``outage postmortem'', and ``failure analysis''. Each candidate page was inspected by a human for apparent relevance before its URL was added to a pool of 659 sources. After removing exact duplicate URLs, we retrieved the publicly available pages programmatically and used Trafilatura to extract relevant text from HTML, producing 343 candidate reports. Content-based de-duplication using Jaccard similarity (threshold $0.8$) reduced this set to 289 reports, from which the LLM filter selected 90 (Sec.~\ref{sec:components}).

\subsection{Filtering}
\label{app:report-filtering}

The incident report filter receives the candidate report text as the user message and uses the following system prompt:

\begin{lstlisting}[style=appendixcode]
Classify the document as exactly one label: GOOD_POSTMORTEM or BAD_POSTMORTEM.

## Strict rule
Label GOOD_POSTMORTEM ONLY if ALL requirements (1-7) below are explicitly satisfied in the text.
If anything is missing, vague, implied, or not tied to concrete evidence, label BAD_POSTMORTEM.

The main goal is to find postmortems that allow us to infer the system structure and write a scenario that explains what happened and why, so that we can use the structure and scenario to write a Python script that generates logs that could have been observed before, during, and after the incident described.

## Requirements

1) Postmortem (post-incident analysis)
Must be a retrospective/postmortem describing an incident after the fact (not just a status update, announcement, changelog, forum thread, Q&A, or article).

2) Cloud/distributed service failure
Must be about a failure of a backend/service hosted in a cloud-based or distributed environment.
Reject if it is mainly: client/desktop issue, purely organizational/process, security-only without operational failure symptoms, single-host/local issue, etc.

3) Explains what happened, why, and observed system behaviors
Must explicitly describe:
- what happened (impact + concrete symptoms),
- why it happened (root cause or causal chain; not speculation-only),
- observed system behaviors during the incident (specific technical behaviors, not generic "degraded").

4) Clear system/component details (names + role)
Must name the relevant system/components AND describe what they do (role/functionality) so the system or parts of the system relevant for the incident can be inferred.
Generic labels like "the service", "backend", "infrastructure" without named components and roles are insufficient.

5) Sufficient detail to reconstruct the scenario over time
Must provide enough detail to reconstruct how the failure unfolded over time:
- either timestamps OR an ordered sequence of events,
AND
- clear progression (detection -> investigation/triage -> mitigation -> recovery) with concrete event descriptions (not just "we fixed it").

6) Log-reflectable incident (explicit incident-time signals)
The problem must be clearly reflectable in system logs/traces (not only business impact).
Must include at least TWO explicit incident-time signals that are specific enough to log, such as:
- explicit error codes/types (e.g., HTTP 5xx/429, gRPC status, DB error, specific exception),
- explicit timeouts/connection failures/retry storms,
- crash/restart/OOM/resource exhaustion with a described symptom,
- failed health checks, leader elections, deployment/rollback failures,
- queue lag/backlog growth, throttling/rate limiting.
If signals are only generic (e.g., "errors occurred", "performance issues") without concrete loggable manifestations, reject.

7) Metrics-based detection (or explicitly metric-convertible)
Must explicitly state EITHER:
- the incident was detected via metrics/monitoring/alerts/anomaly (error rate/latency/saturation/SLO/etc.), OR
- a concrete numeric or categorical signal that is easily and directly convertible to a metric is explicitly mentioned (e.g., "HTTP 500 rate increased", "p95 latency rose", "CPU hit 100%", "queue depth grew", "timeouts spiked").
If detection is only via user reports/support tickets and the text does not explicitly provide a metric-like signal, label BAD_POSTMORTEM.

## Output (JSON only)
Return only:
```json
{
  "label": "GOOD_POSTMORTEM | BAD_POSTMORTEM",
  "missing": [],
  "reasons": [
    "Reason 1: concrete evidence or explicit gap",
    "Reason 2: concrete evidence or explicit gap"
  ]
}
```

## Output field rules
- If label is GOOD_POSTMORTEM: "missing" must be [].
- If label is BAD_POSTMORTEM: "missing" must list every failed check using only:
  ["postmortem", "cloud_distributed", "what_happened", "why",
   "observed_behaviors", "system_structure", "timeline",
   "log_signals", "metrics_signal"]
- "reasons": 1-6 short bullets. Each must quote or tightly paraphrase evidence, or state exactly what is absent. No extra commentary.
\end{lstlisting}

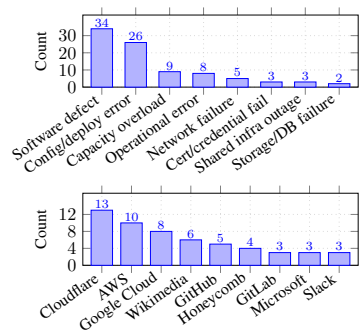
\begin{wrapfigure}[12]{r}{0.35\textwidth}
  \centering
  \vspace{-10pt}
  \resizebox{0.35\textwidth}{!}{\definecolor{catcolor}{RGB}{70, 130, 180}
\definecolor{compcolor}{RGB}{220, 100, 60}
\begin{tikzpicture}
%% ── TOP PLOT: Category Counts ──────────────────────────────────────────────
\begin{axis}[
  name=catplot,
  ybar,
  bar width=12pt,
  width=7cm,
  height=3cm,
  ymin=0, ymax=42,
  enlarge x limits=0.08,
  symbolic x coords={
    {Software defect},
    {Config/deploy error},
    {Capacity overload},
    {Operational error},
    {Network failure},
    {Cert/credential fail},
    {Shared infra outage},
    {Storage/DB failure}
  },
  xtick=data,
  xticklabel style={font=\small, rotate=35, anchor=east},
  ylabel={Count},
  ylabel style={font=\small},
  title style={font=\normalsize, yshift=4pt},
  nodes near coords,
  nodes near coords style={font=\scriptsize, anchor=south, inner sep=1pt},
  every axis plot/.append style={fill=catcolor!80, draw=catcolor!60!black},
  grid=major,
  grid style={dotted, gray!40},
  ytick={0,10,20,30},
]
\addplot coordinates {
  ({Software defect},       34)
  ({Config/deploy error},   26)
  ({Capacity overload},      9)
  ({Operational error},      8)
  ({Network failure},        5)
  ({Cert/credential fail},   3)
  ({Shared infra outage},    3)
  ({Storage/DB failure},     2)
};
\end{axis}
%% ── BOTTOM PLOT: Company Counts (top 8 by count, all ties included) ────────
\begin{axis}[
  name=compplot,
  at={(catplot.below south west)},
  anchor=above north west,
  yshift=0.0cm,
  ybar,
  bar width=12pt,
  width=7cm,
  height=3cm,
  ymin=0, ymax=17,
  enlarge x limits=0.08,
  symbolic x coords={
    {Cloudflare},
    {AWS},
    {Google Cloud},
    {Wikimedia},
    {GitHub},
    {Honeycomb},
    {GitLab},
    {Microsoft},
    {Slack}
  },
  xtick=data,
  xticklabel style={font=\small, rotate=35, anchor=east},
  ylabel={Count},
  ylabel style={font=\small},
  title style={font=\normalsize, yshift=4pt},
  nodes near coords,
  nodes near coords style={font=\scriptsize, anchor=south, inner sep=1pt},
  every axis plot/.append style={fill=compcolor!80, draw=compcolor!60!black},
  grid=major,
  grid style={dotted, gray!40},
  ytick={0,4,8,12},
]
\addplot coordinates {
  ({Cloudflare},     13)
  ({AWS},            10)
  ({Google Cloud},    8)
  ({Wikimedia},       6)
  ({GitHub},          5)
  ({Honeycomb},       4)
  ({GitLab},          3)
  ({Microsoft},       3)
  ({Slack},           3)
};
\end{axis}
\end{tikzpicture}}
  \vspace{-10pt}
  \caption{\textbf{Incident report statistics.} By type (top) and organization (bottom).}
  \label{fig:incident_report_stats}
\end{wrapfigure}
Fig.~\ref{fig:incident_report_stats} illustrates the distribution of failure types and organizations across the 90 curated reports. For organization counts, we drop counts below 3 for clarity.

We see that certain failure types and organizations appear overrepresented. However, manual verification of reports from the top five organizations confirmed the uniqueness of their systems and failure types. We found only one exception (Cloudflare) where a single incident was described twice, first at a high level, then in detail; we retained both due to their substantially different content.

\subsection{Licensing}
\label{app:licensing}

The released artifacts consist of generated incident models, simulators, synthetic log streams, reference answers, benchmark questions, and metadata. We provide the source report URLs for attribution and provenance, but do not redistribute report text, HTML, or images.

We release the benchmark under CC BY 4.0 and the code under Apache 2.0. The benchmark may be used, modified, and redistributed, including commercially, subject to attribution and the license terms. These licenses apply only to the artifacts and code for which we hold the applicable rights; the linked reports remain subject to their publishers' rights and terms.

\section{Incident Model and Simulator Details}
\label{app:incident-model-and-simulator-details}

\subsection{Terminology and incident-model format}
\label{app:incident-model-format}

The \emph{incident model} is an abstraction that can be implemented in different ways, depending on the use case and requirements. We implement it as two YAML files called \texttt{system\_description} and \texttt{scenario}. We chose the YAML format because it could be reliably generated by LLMs and enabled structural verification due to fixed pre-defined schema. These two objects separately define the system description and phased timeline, while overlapping on the emission mechanisms, as defined in Sec.~\ref{sec:components}. Specifically:
\begin{itemize}
    \item 
    \texttt{system\_description}: Defines the static architecture of the system. It outlines the components, connectivity, log templates, background emissions, and request/path flows (detailing how these services interact).
    \item
    \texttt{scenario}: Defines the dynamic timeline of the incident. It details the initial trigger, the chronological sequence of failure events, the observable system behaviors, and the final resolution or mitigation steps.
\end{itemize}

To clarify the terminology mapping between the manuscript text and the released artifacts (codebase and prompts):
\begin{itemize}
    \item 
    \emph{Incident model} (in the paper) $\equiv$ \texttt{system\_description} + \texttt{scenario} (in prompts and as stored model representations) or \texttt{SYSTEM} + \texttt{SCENARIO} (in Python log generation scripts).
    \item
    \emph{Incident simulator} (in the paper) $\equiv$ \emph{log-generation script} (in the prompts).
\end{itemize}

Complete generator and verifier prompts for incident model generation, as well as all system description and scenario objects, can be found in our code and dataset repositories.

\paragraph{Emission mechanisms and parameters.}
Each mechanism is defined by the templates it uses and its timing rule:
\begin{itemize}
    \item \emph{Burst.} A single template is instantiated a specified number of times within a short interval. This models point events such as configuration changes, restarts, or transient error storms. Each \texttt{one\_shots} entry in \texttt{scenario} specifies the template reference (\texttt{ref}), number of messages (\texttt{count}), and emitting hosts (\texttt{hosts}); its associated event determines the burst's timing.
    \item \emph{Background.} A single template is instantiated at a fixed rate for the duration of a phase, either per host or globally across the component's hosts. This models steady-state activity such as health checks, heartbeats, or baseline traffic. In \texttt{system\_description}, the component's \texttt{beh} entries specify the template (\texttt{id}), rate in logs per minute (\texttt{per\_min}), and scope (\texttt{scope}: \texttt{per\_host} or \texttt{global}).
    \item \emph{Path.} An ordered sequence of templates spanning multiple components is instantiated as a unit at a fixed request rate, with consecutive hops separated by sampled latencies. This models a request flow producing correlated logs, which share a trace identifier when tracing is enabled. Each request-flow entry under \texttt{flows} in \texttt{system\_description} specifies the component path (\texttt{path}), ordered template sequence (\texttt{emit}), rate in requests per minute (\texttt{rpm}), and per-hop latency parameters in milliseconds (\texttt{latency\_ms}), given as median and 95th-percentile values.
\end{itemize}
Scenario events select active request flows and scale emission rates and path latencies using \texttt{rate\_multipliers} and \texttt{latency\_multipliers}, respectively. Rates and latency parameters remain fixed until the next event.

Below are the \texttt{system\_description} and \texttt{scenario} YAML schemas with comments specifying the role of different components. Some fields that are not directly related to the incident model specification are omitted for brevity.

\lstdefinelanguage{yaml}{
  keywords={true,false,null,yes,no},
  sensitive=false,
  comment=[l]{\#},
  morestring=[b]',
  morestring=[b]",
}
\lstdefinestyle{code}{
  language=yaml,
  basicstyle=\ttfamily\scriptsize,
  keywordstyle=\color{green!40!black}\bfseries,
  commentstyle=\color{teal}\itshape,
  stringstyle=\color{red!60!black},
  breaklines=true,
  showstringspaces=false,
  columns=fullflexible,
  keepspaces=true,
  tabsize=2,
}

\begin{lstlisting}[style=code]
<system_description>
sys:
  id: <system_name>  # stable incident-model identifier
  desc: <incident-specific system description>
states: {n: normal, f: failure}  # benchmark instances use normal/failure phases

components:
  - id: <component_id>  # machine-readable component identifier
    name: <human-readable name>
    svc: <service-name-or-null>  # service label written into logs
    hosts: [<host1>, ...]  # hostnames available to the simulator
    desc: <component role>
    to:  # directed connectivity used for paths and system diagrams
      - dst: <component_id>
        proto: <https|grpc|kafka|redis|jdbc|tcp|...>
        desc: <edge role>
    logs:  # log templates emitted by the component
      <log_id>:
        lvl: <DEBUG|INFO|WARN|ERROR|CRITICAL>
        msg: "<message template with {variables}>"  # rendered log message
        vars:  # variable types and domains
          <var>: {k: <i|f|ch|uuid|hex|ip|str>, v: <domain>}
        state_vars:  # phase-specific variable domains
          n: {<var>: {k: <type>, v: <normal-domain>}}  # normal phase domains
          f: {<var>: {k: <type>, v: <failure-domain>}} # failure phase domains
    beh:  # background emissions
      n:
        emit:
          - id: <log_id>
            per_min: <float>  # background emission rate
            scope: <per_host|global>
      f:
        emit:
          - id: <log_id>
            per_min: <float>
            scope: <per_host|global>

tracing:
  on: <true|false>
  origins: [<component_id>, ...]
  trace_id: {k: hex, v: 32}

flows:  # path emissions
  n:
    req:
      - id: <normal_flow_id>
        rpm: <requests_per_minute>
        path: [<component_id>, ...]  # logical request path through the system
        emit: [<component_id.log_id>, ...]  # emitted logs along that path
        latency_ms: [[<p50>, <p95>], ...]  # per-hop latency
        trace: <true|false>
  f:
    req:
      - id: <failure_flow_or_variant_id>
        ...
assumptions:
  - <adaptation, traffic, volume, or visibility note>
</system_description>

<scenario>
scenario:
  id: <incident_id>
  title: <brief incident title>
  summary: <normal purpose, trigger, manifestation, impact>
  trigger_desc: <external action or precondition that caused the incident>
  time:
    total_minutes: <2..60>
    phases:
      n: {start_min: 0, end_min: <t>}
      f: {start_min: <t>, end_min: <total_minutes>}
  phases:  # phased timeline
    n:
      flows: [<flow_id>, ...]  # normal request families active before failure
    f:
      events:
        - order: <int>
          at_min: <minute>  # event time; starts a new failure segment
          component: <component_id>
          event: <observed change>
          why: <mechanism and observability>
          flows: [<flow_id>, ...]  # active/requested flow variants after this event
          rate_multipliers: {<flow_or_log_id>: <float>, ...}  # flow IDs or component.log_id keys
          latency_multipliers: {<flow_id>: {p50: <float>, p95: <float>}, ...}
          one_shots:  # burst emissions
            - ref: <component_id.log_id>
              count: <int>
              hosts: [<host>, ...]
</scenario>
\end{lstlisting}

\subsection{High-level fidelity, consistency, and realism of the log streams}
\label{app:incident-model-validity}

While low-level log realism can be formalized and quantified using basic statistical properties and comparisons with open-source production logs (App.~\ref{app:log-stat-comparison}), high-level realism is harder to evaluate due to its ambiguity and lack of ground-truth production log streams with gradations of the properties of interest. We promote such high-level realism by producing log streams based on incident models grounded in real-world incident reports that provide realistic system structures, root causes, failure evolution, and observable evidence, and ensuring the incident models' internal consistency and alignment with their corresponding incident reports. All prompts, along with generation and verification pipelines, are available in our code repo.

\paragraph{Faithfulness and consistency.}
We verify that each incident model preserves the structure and dynamics described in its source report. The LLM verifier receives both the report and the generated model, checking the reported failure mechanism, progression, symptoms, and component roles. It also assesses the plausibility and internal consistency of inferred components and failure dynamics, and whether the model contains enough detail to represent the incident. Algorithmic checks validate connectivity and references to log templates, hosts, and request flows. Acceptance requires ten consecutive verification passes (Sec.~\ref{sec:generation}).

\paragraph{Observability and uncertainty.}
While incident models specify detailed internal system behaviors, adherence to the failure specified in the source report might require that not all aspects of such behaviors can be observed in logs. We represent partial observability by specifying (in the incident model) which internal system behaviors produce observable log evidence. The resulting log stream may leave some causes or mechanisms unresolved. Reference summaries are constrained to the observable effects and must express uncertainty where the evidence does not determine a conclusion. For example, the reference answer in App.~\ref{app:example-answer} identifies database saturation but leaves the cause of the overload unresolved.

\paragraph{Concurrent and distributed behavior.}
The incident model represents distributed incident dynamics through dependencies, concurrent request paths, and phase-dependent behavior across components and hosts. Interleaving path, background, and burst emissions allows normal activity and failure symptoms to coexist in the log stream. For example, the incident model below includes a web dashboard and an API for build agents that share a database. The API continues operating during the dashboard outage, and partial recovery produces successful and failed health checks. These overlapping signals require interpreting observations in the context of the affected components and incident phase.

\paragraph{Scope of realism.}
The incident model determines the range and complexity of behaviors represented in each log stream. We model the minimal subsystem needed to explain each incident, limiting unrelated background noise and incidental production complexity (Sec.~\ref{sec:conclusion}).

\subsection{Concrete incident-model example: \texttt{0b9e6bdf}}

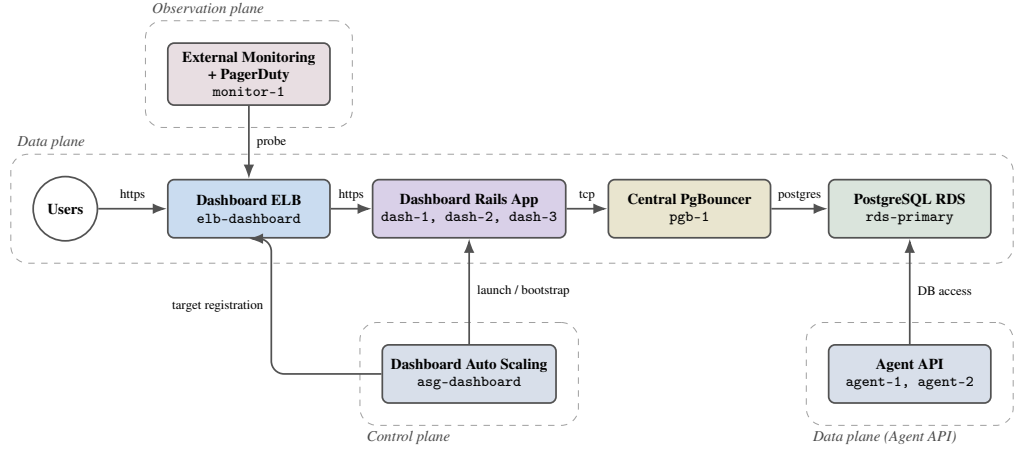
\begin{figure}[!h]
\centering
\resizebox{0.95\linewidth}{!}{%
\begin{tikzpicture}[
  x=1cm,
  y=1cm,
  box/.style={draw=bordergray, rounded corners=3pt, thick, align=center,
              minimum width=2.65cm, minimum height=0.95cm, inner sep=4pt,
              font=\scriptsize, fill=white},
  userstyle/.style={draw=bordergray, circle, thick, align=center,
                    minimum size=1.05cm, inner sep=1pt, font=\scriptsize, fill=white},
  flow/.style={-Latex, thick, draw=bordergray},
  auxflow/.style={-Latex, thick, draw=bordergray, rounded corners=6pt},
  plane/.style={draw=bordergray!55, dashed, rounded corners=8pt, inner sep=10pt},
  planelabel/.style={font=\scriptsize\itshape\color{bordergray!90!black}, inner sep=2pt}
]

% --- Components ---------------------------------------------------------
\node[userstyle]              (user)    at (-3.0, 1.2)  {\textbf{Users}};

\node[box, fill=scorefill]    (monitor) at (0.0, 3.4)   {\textbf{External Monitoring}\\\textbf{+ PagerDuty}\\\texttt{monitor-1}};
\node[box, fill=svcblue]      (elb)     at (0.0, 1.2)   {\textbf{Dashboard ELB}\\\texttt{elb-dashboard}};
\node[box, fill=svclavender]  (app)     at (3.6, 1.2)   {\textbf{Dashboard Rails App}\\\texttt{dash-1, dash-2, dash-3}};
\node[box, fill=answerfill]   (pgb)     at (7.2, 1.2)   {\textbf{Central PgBouncer}\\\texttt{pgb-1}};
\node[box, fill=logfill]      (db)      at (10.8, 1.2)  {\textbf{PostgreSQL RDS}\\\texttt{rds-primary}};
\node[box, fill=questionfill] (asg)     at (3.6, -1.5)  {\textbf{Dashboard Auto Scaling}\\\texttt{asg-dashboard}};
\node[box, fill=questionfill] (agent)   at (10.8, -1.5) {\textbf{Agent API}\\\texttt{agent-1, agent-2}};

% --- Plane backgrounds (behind nodes) -----------------------------------
\begin{pgfonlayer}{background}
  \node[plane, fit=(monitor),
        label={[planelabel, anchor=south west]north west:\,Observation plane}] {};
  \node[plane, fit=(user)(elb)(app)(pgb)(db),
        label={[planelabel, anchor=south west]north west:\,Data plane}] {};
  \node[plane, fit=(asg),
        label={[planelabel, anchor=north west]south west:\,Control plane}] {};
  \node[plane, fit=(agent),
        label={[planelabel, anchor=north west]south west:\,Data plane (Agent API)}] {};
\end{pgfonlayer}

% --- Edges (one arrow per connection) -----------------------------------
\draw[flow]    (user)    -- node[above, font=\tiny] {https}              (elb);
\draw[flow]    (monitor) -- node[right, font=\tiny] {probe}              (elb);
\draw[flow]    (elb)     -- node[above, font=\tiny] {https}              (app);
\draw[flow]    (app)     -- node[above, font=\tiny] {tcp}                (pgb);
\draw[flow]    (pgb)     -- node[above, font=\tiny] {postgres}           (db);
\draw[flow]    (asg)     -- node[right, font=\tiny] {launch / bootstrap} (app);
\draw[auxflow] (asg.west) -- ++(-1.8,0) |- node[pos=0.25, left, font=\tiny] {target registration} (elb.south);
\draw[flow]    (agent)   -- node[right, font=\tiny] {DB access}          (db);

\end{tikzpicture}%
}
\caption{
Compact system topology for incident model \texttt{0b9e6bdf}, annotated by architectural plane.
The dashboard data plane comprises the user and the middle-row components.
The Agent API operates as an independent data-plane workload, sharing only the PostgreSQL
database with the primary dashboard path. Auxiliary systems are positioned off the main
request path, with external monitoring located in the observation plane and the Auto
Scaling Group in the control plane.
}
\label{fig:appendix-0b9e6bdf-diagram}
\end{figure}

Figures~\ref{fig:appendix-0b9e6bdf-diagram} and~\ref{fig:scenario-timeline-log} visualize the system structure and incident progression timeline for the incident model \texttt{0b9e6bdf}. While these figures provide an intuitive, high-level overview of the system state and incident timeline, the YAML excerpt below demonstrates how this data is actually represented and used in our pipeline. For brevity and readability, the full sequence is truncated in this snippet—specifically, intermediate events 2 and 4 have been omitted. This allows us to focus on the core failure mechanisms, their observable effects, and the ultimate resolution steps.

% --- Semantic aliases ------------------------------------------------------
\colorlet{cLogin503}{slsred}
\colorlet{cSlow}{accentteal}
\colorlet{cHCOk}{slsgreen}
\colorlet{cASGLaunch}{anthropicorange}
\colorlet{cRateLimit}{slsblue}
\colorlet{cViewlog503}{openaiblue}
\colorlet{cHCDb}{slsorange}
\colorlet{cHCShard}{answerline}
\colorlet{cASGTerm}{bnodeborder}
\colorlet{cUserdata}{jade}
\colorlet{cRdsConn}{modelgray}
\colorlet{cRdsSlow}{outergray}
\colorlet{cPgbConn}{othergray}
\colorlet{cPdEsc}{bfeedback}
\colorlet{phNormal}{svcsage}
\colorlet{phTrigger}{svcrose}
\colorlet{phProg1}{svcmauve}
\colorlet{phProg2}{answerfill}
\colorlet{phProg3}{svcpeach}
\colorlet{phProg4}{svcblue}
\colorlet{burstRed}{slsred}
\colorlet{burstOrange}{slsorange}
\colorlet{burstGreen}{slsgreen}

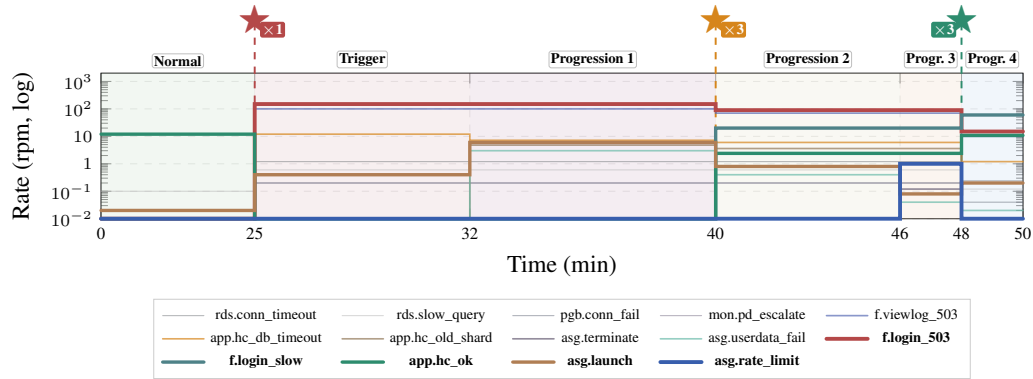
\begin{figure}[!h]
\centering
\resizebox{0.98\linewidth}{!}{%
\begin{tikzpicture}
\begin{semilogyaxis}[
  width=15cm, height=3.7cm,
  xmin=0, xmax=30,
  ymin=0.01, ymax=2000,
  xlabel={Time (min)},
  ylabel={Rate (rpm, log)},
  xtick={0, 5, 12, 20, 26, 28, 30},
  xticklabels={0, 25, 32, 40, 46, 48, 50},
  xticklabel style={font=\scriptsize},
  yticklabel style={font=\scriptsize},
  ytick={0.01, 0.1, 1, 10, 100, 1000},
  yticklabels={$10^{-2}$, $10^{-1}$, $1$, $10$, $10^{2}$, $10^{3}$},
  xlabel style={font=\scriptsize, yshift=2pt},
  ylabel style={font=\scriptsize},
  xlabel near ticks,
  ylabel near ticks,
  legend style={
    at={(0.5, -0.55)}, anchor=north,
    legend columns=5, font=\tiny,
    draw=gray!40,
    /tikz/every even column/.append style={column sep=0.2cm},
  },
grid=major,
major grid style={gray!20, dashed},
  axis on top=false,
  clip=false,
  enlarge x limits=false,
  every axis plot/.style={mark=none, line cap=round},
]

% --- Phase background bands (highly transparent) ---------------------------
\fill[phNormal,  opacity=0.25] (axis cs:0,0.01)  rectangle (axis cs:5,2000);
\fill[phTrigger, opacity=0.25] (axis cs:5,0.01)  rectangle (axis cs:12,2000);
\fill[phProg1,   opacity=0.25] (axis cs:12,0.01) rectangle (axis cs:20,2000);
\fill[phProg2,   opacity=0.25] (axis cs:20,0.01) rectangle (axis cs:26,2000);
\fill[phProg3,   opacity=0.25] (axis cs:26,0.01) rectangle (axis cs:28,2000);
\fill[phProg4,   opacity=0.25] (axis cs:28,0.01) rectangle (axis cs:30,2000);

% --- Vertical burst marker lines (extend through annotation stack) ---------
\draw[burstRed,    dashed, thick] (axis cs:5,0.01)  -- ([yshift=24pt] axis cs:5,2000);
\draw[burstOrange, dashed, thick] (axis cs:20,0.01) -- ([yshift=24pt] axis cs:20,2000);
\draw[burstGreen,  dashed, thick] (axis cs:28,0.01) -- ([yshift=24pt] axis cs:28,2000);

% --- Background metrics (lowest tier: thin, faint, drawn first) ------------
\addplot[const plot, line width=0.45pt, color=cRdsConn, opacity=0.55]
  coordinates { (0,0.01) (5,1.2) (28,0.24) (30,0.24) };
\addlegendentry{rds.conn\_timeout}

\addplot[const plot, line width=0.45pt, color=cRdsSlow, opacity=0.55]
  coordinates { (0,0.1) (5,0.6) (28,0.12) (30,0.12) };
\addlegendentry{rds.slow\_query}

\addplot[const plot, line width=0.45pt, color=cPgbConn, opacity=0.55]
  coordinates { (0,0.01) (5,0.2) (28,0.04) (30,0.04) };
\addlegendentry{pgb.conn\_fail}

\addplot[const plot, line width=0.45pt, color=cPdEsc, opacity=0.55]
  coordinates { (0,0.01) (5,0.2) (30,0.2) };
\addlegendentry{mon.pd\_escalate}

% --- Secondary lines (middle tier) -----------------------------------------
\addplot[const plot, line width=0.7pt, color=cViewlog503, opacity=0.7]
  coordinates { (0,0.01) (5,100) (20,70) (28,10) (30,10) };
\addlegendentry{f.viewlog\_503}

\addplot[const plot, line width=0.7pt, color=cHCDb, opacity=0.7]
  coordinates { (0,0.01) (5,12) (12,7.2) (20,6) (28,1.2) (30,1.2) };
\addlegendentry{app.hc\_db\_timeout}

\addplot[const plot, line width=0.7pt, color=cHCShard, opacity=0.7]
  coordinates { (0,0.01) (12,4.8) (20,3.6) (28,0.01) (30,0.01) };
\addlegendentry{app.hc\_old\_shard}

\addplot[const plot, line width=0.7pt, color=cASGTerm, opacity=0.7]
  coordinates { (0,0.02) (5,0.4) (12,6) (20,0.8) (26,0.12) (28,0.2) (30,0.2) };
\addlegendentry{asg.terminate}

\addplot[const plot, line width=0.7pt, color=cUserdata, opacity=0.7]
  coordinates { (0,0.01) (12,3) (20,0.4) (26,0.04) (28,0.02) (30,0.02) };
\addlegendentry{asg.userdata\_fail}

% --- Hero series (drawn last, on top) --------------------------------------
\addplot[const plot, line width=1.6pt, color=cLogin503]
  coordinates { (0,0.01) (5,150) (20,90) (28,15) (30,15) };
\addlegendentry{\textbf{f.login\_503}}

\addplot[const plot, line width=1.4pt, color=cSlow]
  coordinates { (0,0.01) (20,20) (28,60) (30,60) };
\addlegendentry{\textbf{f.login\_slow}}

\addplot[const plot, line width=1.4pt, color=cHCOk]
  coordinates { (0,12) (5,0.01) (20,2.4) (28,10.8) (30,10.8) };
\addlegendentry{\textbf{app.hc\_ok}}

\addplot[const plot, line width=1.4pt, color=cASGLaunch]
  coordinates { (0,0.02) (5,0.4) (12,6) (20,0.8) (26,0.08) (28,0.2) (30,0.2) };
\addlegendentry{\textbf{asg.launch}}

\addplot[const plot, line width=1.6pt, color=cRateLimit]
  coordinates { (0,0.01) (26,1.0) (28,0.01) (30,0.01) };
\addlegendentry{\textbf{asg.rate\_limit}}

% --- Phase labels (positioned with pixel yshift above plot top) ------------
\node[font=\tiny\bfseries, fill=white, inner sep=0.8pt, draw=gray!50,
      rounded corners=1pt, anchor=south, yshift=2pt]
  at (axis cs:2.5, 2000) {Normal};
\node[font=\tiny\bfseries, fill=white, inner sep=0.8pt, draw=gray!50,
      rounded corners=1pt, anchor=south, yshift=2pt]
  at (axis cs:8.5, 2000) {Trigger};
\node[font=\tiny\bfseries, fill=white, inner sep=0.8pt, draw=gray!50,
      rounded corners=1pt, anchor=south, yshift=2pt]
  at (axis cs:16, 2000) {Progression~1};
\node[font=\tiny\bfseries, fill=white, inner sep=0.8pt, draw=gray!50,
      rounded corners=1pt, anchor=south, yshift=2pt]
  at (axis cs:23, 2000) {Progression~2};
\node[font=\tiny\bfseries, fill=white, inner sep=0.8pt, draw=gray!50,
      rounded corners=1pt, anchor=south, yshift=2pt]
  at (axis cs:27, 2000) {Progr.~3};
\node[font=\tiny\bfseries, fill=white, inner sep=0.8pt, draw=gray!50,
      rounded corners=1pt, anchor=south, yshift=2pt]
  at (axis cs:29, 2000) {Progr.~4};

% --- Burst stars + count badges (yshift higher than phase labels) ----------
\node[color=burstRed, font=\Large, anchor=south, yshift=12pt]
  at (axis cs:5, 2000) {$\bigstar$};
\node[font=\tiny\bfseries, fill=burstRed, text=white, inner sep=0.8pt,
      rounded corners=1pt, anchor=west, xshift=2pt, yshift=18pt]
  at (axis cs:5, 2000) {$\times$1};

\node[color=burstOrange, font=\Large, anchor=south, yshift=12pt]
  at (axis cs:20, 2000) {$\bigstar$};
\node[font=\tiny\bfseries, fill=burstOrange, text=white, inner sep=0.8pt,
      rounded corners=1pt, anchor=west, xshift=2pt, yshift=18pt]
  at (axis cs:20, 2000) {$\times$3};

\node[color=burstGreen, font=\Large, anchor=south, yshift=12pt]
  at (axis cs:28, 2000) {$\bigstar$};
\node[font=\tiny\bfseries, fill=burstGreen, text=white, inner sep=0.8pt,
      rounded corners=1pt, anchor=east, xshift=-2pt, yshift=18pt]
  at (axis cs:28, 2000) {$\times$3};

\end{semilogyaxis}
\end{tikzpicture}%
}
\caption{Compact timeline of incident model \texttt{0b9e6bdf}. The figure illustrates the normal phase and all five ordered events of the 50-minute scenario.
Thick lines highlight the dominant incident trajectory, thin lines represent
secondary background activities, and stars denote one-shot log bursts at events 1, 3,
and 5. Log and flow IDs are abbreviated for brevity.}
\label{fig:scenario-timeline-log}
\end{figure}

The YAML excerpt below shows the \texttt{system\_description} and \texttt{scenario} for the incident model \texttt{0b9e6bdf}. Some logs, flows, and fields are omitted for brevity.

\begin{lstlisting}[style=appendixcode]
sys:
  id: buildkite_dashboard_2016
  desc: >
    Hosted CI service with a web dashboard behind an ELB, Rails app instances,
    central PgBouncer, and PostgreSQL. The dashboard degrades, but the Agent API
    continues polling throughout the incident.

states: {n: normal, f: failure}

components:
  - id: dashboard_elb
    name: Dashboard ELB
    hosts: [elb-dashboard]
    to: [{dst: dashboard_app, proto: https}]
    logs:
      req_no_healthy:
        lvl: WARN
        msg: "ELB request {method} {url_path} -> 503 no_healthy_backends latency_ms={lat_ms} req={req_id}"
      hc_result_500:
        lvl: WARN
        msg: "ELB healthcheck result target={target} status=500 latency_ms={lat_ms} req={req_id}"

  - id: dashboard_app
    name: Dashboard Rails App
    hosts: [dash-1, dash-2, dash-3]
    to:
      - {dst: pgbouncer_central, proto: tcp}
      - {dst: dashboard_elb, proto: https}
    logs:
      req_started:
        lvl: INFO
        msg: "Started {method} {url_path} controller={controller} action={action} req={req_id}"
      healthcheck_db_timeout:
        lvl: ERROR
        msg: "Healthcheck failed: database ping timeout waited_ms={waited_ms} shard=two req={req_id}"
      healthcheck_old_shard_error:
        lvl: ERROR
        msg: "Healthcheck failed: database shard missing shard=one err=\"{err}\" req={req_id}"

  - id: pgbouncer_central
    name: Central PgBouncer
    hosts: [pgb-1]
    to: [{dst: postgres_rds, proto: postgres}]
    logs:
      pool_stats:
        lvl: INFO
        msg: "pgbouncer stats db=buildkite active={active} waiting={waiting} max_client_conn={max_conn} avg_wait_ms={avg_wait_ms}"

  - id: postgres_rds
    name: PostgreSQL RDS
    hosts: [rds-primary]
    to:
      - {dst: pgbouncer_central, proto: postgres}
      - {dst: dashboard_app, proto: postgres}
      - {dst: agent_api, proto: postgres}
    logs:
      rds_metrics:
        lvl: INFO
        msg: "rds metrics cpu_pct={cpu_pct} conn_used={conn_used} qps={qps} profile={profile}"

  - id: agent_api
    name: Agent API
    hosts: [agent-1, agent-2]
    to: [{dst: postgres_rds, proto: postgres}]
    logs:
      agent_req:
        lvl: INFO
        msg: "agent-api {method} {url_path} -> {status} dur_ms={dur_ms} req={req_id}"

  - id: autoscaling
    name: Dashboard Auto Scaling
    hosts: [asg-dashboard]
    to:
      - {dst: dashboard_elb, proto: aws-api}
      - {dst: dashboard_app, proto: ec2-bootstrap}
    logs:
      userdata_fetch_rev_failed:
        lvl: ERROR
        msg: "userdata fetch deploy revision failed url=https://buildkite.com/_secret/version err=\"{err}\""
      ami_rollout_started:
        lvl: INFO
        msg: "ami rollout started ami={ami_id} revision_source={source} reason=\"{reason}\""

  - id: monitoring
    name: External Monitoring + PagerDuty
    hosts: [monitor-1]
    to: [{dst: dashboard_elb, proto: https}]
    logs:
      pagerduty_incident:
        lvl: ERROR
        msg: "pagerduty incident opened service=dashboard severity=high incident_id={incident_id}"

flows:
  n:
    req:
      - id: dashboard_login
        rpm: 120.0
        path: [dashboard_elb, dashboard_app, pgbouncer_central,
               postgres_rds, dashboard_app, dashboard_elb]
        emit: [dashboard_app.req_started,
               dashboard_app.req_completed_ok,
               dashboard_elb.req_ok]
        latency_ms: [[1, 10], [30, 800], [1, 20]]
      - id: dashboard_elb_healthcheck
        rpm: 12.0
        path: [dashboard_elb, dashboard_app, dashboard_elb]
        emit: [dashboard_elb.hc_access,
               dashboard_app.healthcheck_ok,
               dashboard_elb.hc_result_ok]
      - id: agent_job_poll
        rpm: 300.0
        path: [agent_api, postgres_rds]
        emit: [agent_api.agent_req]

  f:
    req:
      - id: dashboard_login_503
        rpm: 150.0
        path: [dashboard_elb]
        emit: [dashboard_elb.req_no_healthy]
      - id: dashboard_elb_healthcheck_fail_db
        rpm: 12.0
        path: [dashboard_elb, dashboard_app, dashboard_elb]
        emit: [dashboard_elb.hc_access,
               dashboard_app.healthcheck_db_timeout,
               dashboard_elb.hc_result_500]
      - id: dashboard_elb_healthcheck_fail_old_shard
        rpm: 12.0
        path: [dashboard_elb, dashboard_app, dashboard_elb]
        emit: [dashboard_elb.hc_access,
               dashboard_app.healthcheck_old_shard_error,
               dashboard_elb.hc_result_500]

scenario:
  id: buildkite_outage_aug_2016_compressed
  time:
    total_minutes: 50
    phases:
      n: {start_min: 0, end_min: 25}
      f: {start_min: 25, end_min: 50}
  phases:
    f:
      events:
        - order: 1
          at_min: 25
          component: postgres_rds
          event: >
            PostgreSQL saturation begins; DB-dependent dashboard health checks fail,
            and the ELB starts returning 503 because targets are drained.
          flows: [dashboard_elb_healthcheck_fail_db, dashboard_login_503,
                  dashboard_view_build_log_503, agent_job_poll]
          rate_multipliers:
            dashboard_login_slow: 0.0
            dashboard_elb_healthcheck_fail_old_shard: 0.0
            dashboard_elb_healthcheck_ok: 0.0
            autoscaling.userdata_fetch_rev_failed: 0.0
            autoscaling.launch_failed_rate_limit: 0.0
            postgres_rds.rds_metrics_recovered: 0.0
            pgbouncer_central.pool_stats_recovered: 0.0
          latency_multipliers:
            dashboard_elb_healthcheck_fail_db: {p50: 1.2, p95: 1.2}
            agent_job_poll: {p50: 1.1, p95: 1.2}
          one_shots:
            - {ref: monitoring.pagerduty_incident, count: 1, hosts: [monitor-1]}

        - order: 3
          at_min: 40
          component: autoscaling
          event: >
            Operators lock desired capacity and register borrowed instances to the ELB;
            some health checks recover, but many dashboard requests still fail or run slowly.
          flows: [dashboard_elb_healthcheck_ok, dashboard_elb_healthcheck_fail_db,
                  dashboard_elb_healthcheck_fail_old_shard, dashboard_login_slow,
                  dashboard_login_503, dashboard_view_build_log_503]
          rate_multipliers:
            dashboard_login_503: 0.6
            dashboard_view_build_log_503: 0.7
            dashboard_login_slow: 1.0
            dashboard_elb_healthcheck_fail_db: 0.5
            dashboard_elb_healthcheck_fail_old_shard: 0.3
            dashboard_elb_healthcheck_ok: 0.2
            autoscaling.instance_launch: 2.0
            autoscaling.instance_terminate: 2.0
            autoscaling.userdata_fetch_rev_failed: 2.0
          latency_multipliers:
            dashboard_login_slow: {p50: 1.2, p95: 1.3}
          one_shots:
            - {ref: autoscaling.desired_capacity_set, count: 1, hosts: [asg-dashboard]}
            - {ref: autoscaling.register_instance_to_elb, count: 2, hosts: [asg-dashboard]}

        - order: 5
          at_min: 48
          component: postgres_rds
          event: >
            Operators upgrade RDS, roll out a fixed AMI bootstrap path, and roll back
            pooling topology; health checks mostly recover and 503s drop sharply.
          flows: [dashboard_elb_healthcheck_ok, dashboard_elb_healthcheck_fail_db,
                  dashboard_login_slow, dashboard_login_503, dashboard_view_build_log_503]
          rate_multipliers:
            dashboard_elb_healthcheck_ok: 0.9
            dashboard_elb_healthcheck_fail_db: 0.1
            dashboard_elb_healthcheck_fail_old_shard: 0.0
            dashboard_login_503: 0.1
            dashboard_view_build_log_503: 0.1
            dashboard_login_slow: 3.0
            postgres_rds.rds_metrics: 0.0
            postgres_rds.rds_metrics_recovered: 1.0
            postgres_rds.connection_timeout: 0.2
            postgres_rds.slow_query: 0.2
            pgbouncer_central.pool_stats: 0.0
            pgbouncer_central.pool_stats_recovered: 1.0
            pgbouncer_central.server_connect_failed: 0.2
            autoscaling.userdata_fetch_rev_failed: 0.1
            autoscaling.instance_launch: 0.5
            autoscaling.instance_terminate: 0.5
            autoscaling.launch_failed_rate_limit: 0.2
          latency_multipliers:
            dashboard_login_slow: {p50: 0.7, p95: 0.6}
            agent_job_poll: {p50: 0.95, p95: 0.95}
          one_shots:
            - {ref: postgres_rds.modify_instance, count: 1, hosts: [rds-primary]}
            - {ref: autoscaling.ami_rollout_started, count: 1, hosts: [asg-dashboard]}
            - {ref: pgbouncer_central.config_change, count: 1, hosts: [pgb-1]}
\end{lstlisting}

\subsection{Incident model statistics}
\label{app:incident-model-stats}

Table~\ref{tab:incident-model-stats} presents the summary statistics for the generated incident models. We deliberately design our generation prompts to target a compact incident-specific model that is small enough for reliable generation and verification by the LLM, yet sufficient to capture realistic system structures and incident progression. As detailed in the table, the generated models exhibit significant variance and scale, with some instances reaching up to 211 distinct hosts and 180 variable slots, demonstrating the robust generation capabilities of our pipeline. The median accepted model has six components, 27 log templates, eight request-flow variants, and four failure events. 

\begin{table}[h]
  \centering
  \small
  \caption{Structural statistics of the 90 accepted incident models.}
  \label{tab:incident-model-stats}
  \begin{tabular}{lrrrr}
    \toprule
    Quantity & Mean & SD & Median & Range \\
    \midrule
    Components & 6.11 & 1.45 & 6.0 & 3--10 \\
    Hosts & 16.38 & 23.66 & 12.5 & 7--211 \\
    Component edges & 7.91 & 2.90 & 7.0 & 3--16 \\
    Log templates & 26.96 & 7.85 & 27.0 & 12--46 \\
    Variable slots & 81.84 & 25.28 & 78.0 & 42--180 \\
    Flow variants & 8.34 & 3.44 & 8.0 & 3--18 \\
    Normal flow variants & 2.76 & 1.34 & 3.0 & 1--7 \\
    Failure flow variants & 5.59 & 2.32 & 5.0 & 2--11 \\
    Background emission sources & 14.28 & 5.54 & 14.0 & 5--28 \\
    Failure events & 4.22 & 1.13 & 4.0 & 2--7 \\
    Total scenario minutes & 48.21 & 8.87 & 50.0 & 24--60 \\
    Normal-phase minutes & 22.04 & 4.06 & 21.5 & 12--30 \\
    Failure-phase minutes & 26.17 & 6.48 & 25.0 & 12--45 \\
    \bottomrule
  \end{tabular}
\end{table}

\subsection{Simulator example and generated logs}
\label{app:simulator-example}

For each accepted incident model, we generate a deterministic Python script that embeds the model and converts the scenario events into piecewise-constant failure segments. Within these segments, the script instantiates request-flow families, emits background logs and one-shot events, sorts all rows chronologically, and outputs a standard CSV format (\texttt{timestamp}, \texttt{level}, \texttt{message}, \texttt{trace\_id}, \texttt{service}, \texttt{host}). To give a representative example, executing the simulator for incident \texttt{0b9e6bdf} produces 43368 log message lines. The following excerpt summarizes the core logic: it derives failure segments from the scenario, applies rate and latency controls, and emits background logs, one-shot event markers, and request-flow instances inside each segment.

\begin{lstlisting}[style=appendixcode]
def derive_failure_segments() -> List[Dict[str, Any]]:
    fstart = SCENARIO["scenario"]["time"]["phases"]["f"]["start_min"]
    fend = SCENARIO["scenario"]["time"]["phases"]["f"]["end_min"]
    events = sorted(SCENARIO["scenario"]["phases"]["f"]["events"],
                    key=lambda e: (e["at_min"], e["order"]))

    boundaries = sorted(set([fstart] + [e["at_min"] for e in events] + [fend]))
    rate_flow, rate_bg, lat_flow = {}, {}, {}
    segs = []
    events_by_min: Dict[int, List[Dict[str, Any]]] = {}
    for e in events:
        events_by_min.setdefault(int(e["at_min"]), []).append(e)

    for s, e in zip(boundaries[:-1], boundaries[1:]):
        for ev in events_by_min.get(int(s), []):
            for k, m in (ev.get("rate_multipliers") or {}).items():
                (rate_bg if "." in k else rate_flow)[k] = float(m)
            for fk, mult in (ev.get("latency_multipliers") or {}).items():
                lat_flow[fk] = {"p50": float(mult.get("p50", 1.0)),
                                "p95": float(mult.get("p95", 1.0))}
        segs.append({
            "start_min": int(s),
            "end_min": int(e),
            "rate_flow": dict(rate_flow),
            "rate_bg": dict(rate_bg),
            "lat_flow": dict(lat_flow),
        })
    return segs

FAILURE_SEGS = derive_failure_segments()

def simulate() -> pd.DataFrame:
    base_time = datetime(2016, 8, 10, 12, 0, 0, tzinfo=timezone.utc)
    rows: List[LogRow] = []

    # Normal background emissions and normal flows.
    ...

    # Failure segments: one-shots, background, then flow instances.
    for seg in FAILURE_SEGS:
        seg_start_dt = base_time + timedelta(minutes=seg["start_min"])
        seg_end_dt = base_time + timedelta(minutes=seg["end_min"])

        for os in seg.get("one_shots", []):
            emit_log(rows, ..., os["ref"], "f", bound={}, host_override=host)

        for cid, cinfo in COMP.items():
            for src in cinfo.get("beh", {}).get("f", {}).get("emit", []):
                mult = get_rate_multiplier_bg(seg, f"{cid}.{src['id']}")
                ...
                emit_log(rows, t, f"{cid}.{src['id']}", "f", bound={}, host_override=host)

        for fdef in SYSTEM["flows"]["f"]["req"]:
            flow_id = fdef["id"]
            eff_rpm = float(fdef["rpm"]) * get_rate_multiplier_flow(seg, flow_id)
            ...
            lat_mult = get_flow_latency_multiplier(seg, flow_id)
            simulate_flow_instance(rows, "f", fdef, st, lat_mult, inst_key)

    df = pd.DataFrame(...).sort_values("timestamp", kind="mergesort")
    df["timestamp"] = df["timestamp"].apply(iso_z)
    return df[["timestamp", "level", "message", "trace_id", "service", "host"]]
\end{lstlisting}

The generated logs provide a concrete realization of the incident model, translating the abstract topology and timeline into an observable diagnostic artifact. The short excerpt below shows normal operation, the onset of the dashboard outage, the later bootstrap failure on replacement instances, and recovery attempts near the end of the window.

\begin{lstlisting}[style=appendixcode]
timestamp,level,message,trace_id,service,host
2016-08-10T12:00:00.213Z,INFO,Started GET /login controller=sessions action=new req=099e73940896feb9,,buildkite-web,dash-2
2016-08-10T12:00:00.243Z,INFO,Completed 200 in 30ms req=099e73940896feb9,,buildkite-web,dash-2
2016-08-10T12:00:00.282Z,INFO,ELB request GET /login -> 200 target=dash-2 latency_ms=70 req=099e73940896feb9,,aws-elb,elb-dashboard
2016-08-10T12:25:00.686Z,ERROR,pagerduty incident opened service=dashboard severity=high incident_id=2722cd23,,monitoring,monitor-1
2016-08-10T12:25:02.641Z,ERROR,Healthcheck failed: database ping timeout waited_ms=200 shard=two req=3390cb390814cd36,,buildkite-web,dash-1
2016-08-10T12:25:02.685Z,WARN,ELB healthcheck result target=dash-1 status=500 latency_ms=244 req=3390cb390814cd36,,aws-elb,elb-dashboard
2016-08-10T12:25:03.027Z,WARN,ELB request GET /login -> 503 no_healthy_backends latency_ms=15 req=81d5a8c0b5b71350,,aws-elb,elb-dashboard
2016-08-10T12:32:09.807Z,ERROR,"userdata fetch deploy revision failed url=https://buildkite.com/_secret/version err=\"connection refused\"",,aws-asg,asg-dashboard
2016-08-10T12:32:19.060Z,ERROR,"Healthcheck failed: database shard missing shard=one err=\"relation does not exist\" req=344f67da016bb76d",,buildkite-web,dash-3
2016-08-10T12:48:00.459Z,INFO,"pgbouncer config change pooling_mode=per_host reason=\"rollback_from_central\"",,pgbouncer,pgb-1
2016-08-10T12:48:00.472Z,INFO,"ami rollout started ami=i-a5f658ea revision_source=s3_file reason=\"fix_bootstrap_revision_source\"",,aws-asg,asg-dashboard
\end{lstlisting}

% These rows illustrate the evidence chain induced by the scenario: database and pooling telemetry establish resource pressure, health-check logs explain why ELB targets are drained, user-facing 503s expose impact, Auto Scaling logs show why replacement capacity is unstable, and one-shot mitigation markers indicate when recovery actions enter the timeline.

\section{Log-Statistics Comparison Details}
\label{app:log-stat-comparison}
\begin{figure}[h]
  \centering
  \includegraphics[width=\linewidth]{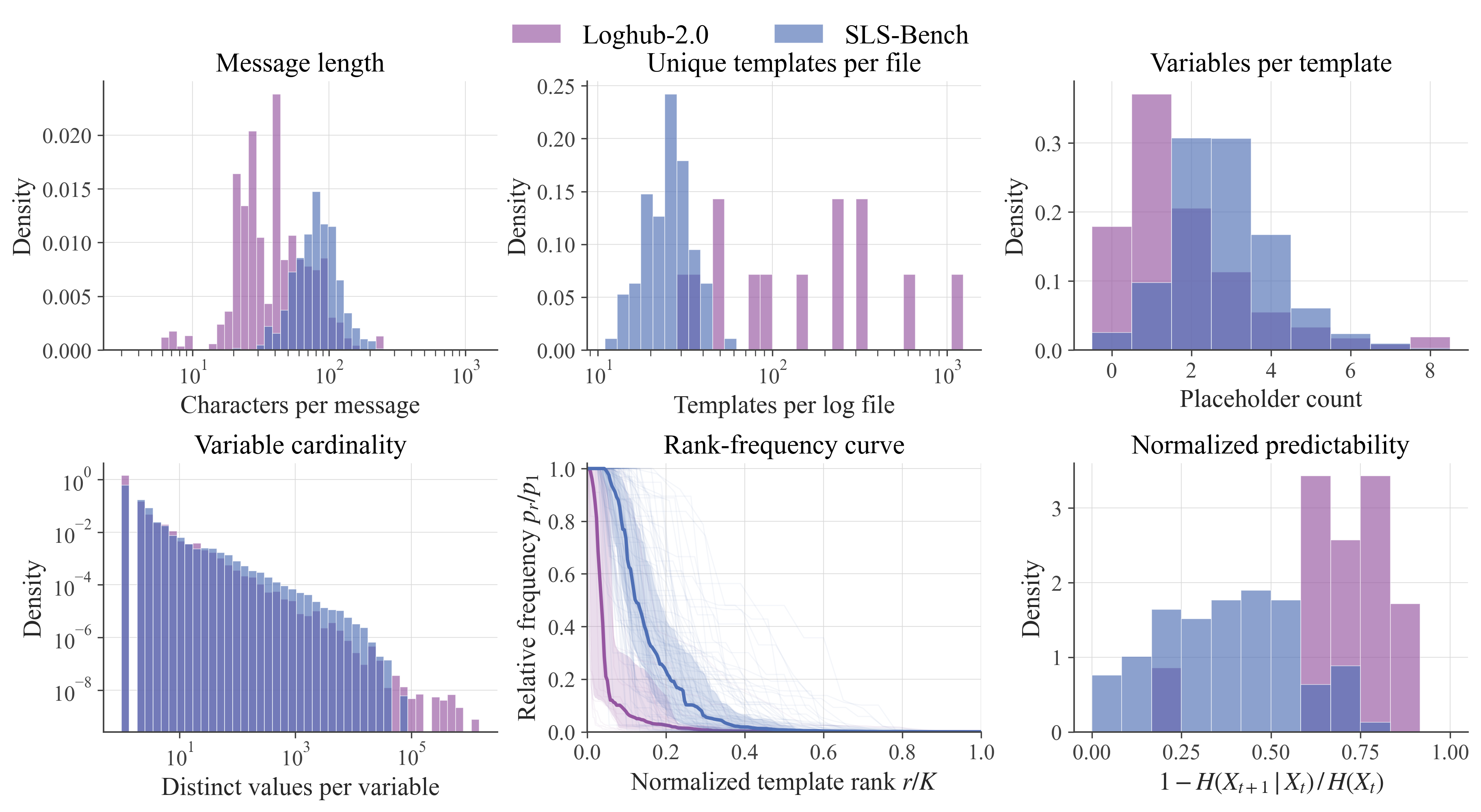}
  \caption{Distributional comparison between Loghub-2.0 and \slsbench. Panels show message length, unique templates per file, variables per template, variable cardinality, normalized template rank--frequency curves, and adjacent-template predictability.}
  \label{fig:loghub-vs-sls-histograms}
\end{figure}
We compare \slsbench\ logs with Loghub-2.0 log files from the Hadoop, HDFS, OpenStack, Spark, Zookeeper, BGL, Thunderbird, HPC, Proxifier, Apache, OpenSSH, HealthApp, Mac, and Linux datasets.
% Several statistics are file-size dependent: a longer file can expose more templates, more values for the same template variables, and different empirical transition frequencies. To keep the comparison at the same scale as our benchmark logs, we truncate each Loghub-2.0 file to its first 100{,}000 rows before computing the statistics. This preserves chronological order while matching the maximum \slsbench\ stream length.
For Loghub-2.0, templates are the released \texttt{EventTemplate} annotations. For \slsbench, templates come from the incident-model system descriptions.
% Each \slsbench\ template contains named slots, which we normalize to \texttt{<*>} for comparison. Template matching is constrained by the row's \texttt{host} and \texttt{service}: a row is matched only against templates emitted by the corresponding component. When more than one candidate template matches a message, we choose the most specific match, preferring the template with more literal text and then the one with fewer slots.

Fig.~\ref{fig:loghub-vs-sls-histograms} shows six distributions. \emph{Message length} is the number of characters in the emitted message text. \emph{Unique templates per file} counts the distinct observed templates in each log stream. \emph{Variables per template} counts the number of placeholder slots in each observed template. \emph{Variable cardinality} counts the number of distinct observed values for each template slot within a file. 

The \emph{rank--frequency} panel summarizes how log volume is distributed across templates. For each file, we sort templates by descending frequency and plot \((r/K, p_r/p_1)\), where \(K\) is the number of observed templates, \(r\) is the one-indexed rank, \(p_r\) is the empirical share of the template at rank \(r\), and \(p_1\) is the share of the most frequent template. The panel overlays the per-file curves and the dataset-level median with their interquartile range band. Adjacent-template \emph{normalized predictability} measures local stream regularity using the observed template sequence: \(1-H(X_{t+1}\mid X_t)/H(X_t)\), where \(X_t\) is the template at position \(t\) and \(H\) is Shannon entropy. Values near one indicate that the next template is almost determined by the current template; values near zero indicate weak one-step predictability.

Comparing the distributions for the two datasets reveals the following similarities and differences:
\begin{itemize}
    \item The message length distributions overlap significantly, with \slsbench\ being narrower and on average longer. In part, this is due to our log messages containing more variables and often including various identifiers such as trace or request IDs that are typically long strings of random characters.
    \item The most notable difference is the number of unique templates. While for some Loghub-2.0 systems the number of templates overlaps with \slsbench, there are many cases with hundreds of unique templates (notably Mac and Linux systems that have 626 and 338 templates, respectively). This discrepancy is expected as a consequence of our design choice to infer only the minimal relevant system subset that represents and explains the incident. This prevents our incident model from having multiple components unrelated to the failure and contributing their templates to the log stream. This aspect can be easily adjusted by injecting real or synthetic log templates unrelated to the incident into the log stream to reduce the signal-to-noise ratio.
    \item The number of variables, while being slightly larger on average for our logs, significantly overlaps with Loghub-2.0, indicating realistic log template structure. Similarly, variable cardinality distributions have very similar shapes across the datasets, indicating that, similarly to real logs, the majority of variables are only instantiated with a few unique values. Only a small fraction has a large number of unique values.
    \item Another notable difference is in how templates are used in real and our synthetic logs. The rank--frequency curves show that both \slsbench\ and Loghub-2.0 have distinct heavy-tailed curves, indicating that there is a relatively small fraction of templates that are used most often, and a long tail of templates that are used very rarely. Real system logs, however, demonstrate a notably steeper drop, indicating that a few templates dominate the real-world log streams. \slsbench, on the other hand, shows larger template utilization without extreme skewness towards a small fraction of dominant templates. These results are consistent with the previous result on the number of unique templates in both datasets.
    \item Finally, we compare the local structure of \slsbench\ and Loghub-2.0 in terms of normalized predictability. We see that real log streams tend to be much more predictable than \slsbench\ log messages. This is consistent with rank--frequency results where a few dominant templates occupy most of the log stream, making it more monotone and predictable. Our log messages, on the other hand, overlap bursts, background, and path emissions without repeating the same log template multiple times, making our synthetic log streams less locally predictable.
\end{itemize}

\section{Fixed Question and Example Reference Answer}
\label{app:fixed-question-and-example-answer}

Each benchmark problem uses the same question shown in App.~\ref{app:fixed-question}. For problem \texttt{0b9e6bdf}, the answer is given in App.~\ref{app:example-answer}; it is shortened to keep the appendix compact, while the released benchmark files contain the full reference answer.

\subsection{Fixed question}
\label{app:fixed-question}

\begin{lstlisting}[style=appendixtext]
Analyze the provided log file and write an incident investigation report.

Use the logs to reconstruct what system this is, how it normally behaves, when the behavior changes, and the single best diagnosis of the incident. Base the report on evidence visible in the logs plus reasonable domain knowledge. When the logs do not fully determine a conclusion, state the best-supported interpretation and make the uncertainty explicit.

Use exactly this structure:

## Incident synopsis
Give a short bullet list with one bullet per phase in chronological order. Each bullet should name the phase and summarize its main observable state or change in one sentence.

## Reconstructed system
Use exactly these `###` subsections in this order:
- Inferred purpose
- Main services/components
- Main interactions

Describe the system only to the extent the logs support it.
Prioritize the core request or data path. Mention monitoring, probes, operator tooling, deployment or restore tooling, and other support-plane actors only when they are diagnostically central.
Prefer names and roles that are visible in service names, hostnames, trace-linked interactions, or message text. Do not introduce hidden component labels or concrete external dependencies unless the logs support them.
When a role, dependency, or interaction is inferred rather than explicit, use appropriately hedged language such as `appears to`, `likely`, or `suggests`.

## Timeline
Partition the full log into phases with [inclusive, exclusive) time ranges in minutes from the start of the file. Include the normal baseline, the trigger phase, and each subsequent failure-progression phase. Compare each phase primarily to the immediately preceding phase, but you may also refer explicitly to earlier phases when that helps explain the current one.

Use one `###` heading per phase in chronological order:
- `### Baseline -- [start, end)`
- `### Trigger -- [start, end)`
- `### Progression 1 -- [start, end)`
- `### Progression 2 -- [start, end)`
- and so on as needed

Under each phase heading, use exactly these `####` subsections in this order:
- Summary
- Evidence
- Reasoning

For each phase:
- `Summary` should state the main phase-level claim in 2--3 sentences. For non-baseline phases, make the main visible change relative to the immediately preceding phase clear. Do not turn this subsection into an evidence dump.
- `Evidence` should stay anchored to the log surface. Cite the concrete evidence that supports the phase reading: representative log templates with "..." placeholders, request or trace-linked sequences, retry patterns, bursty operational markers, service/host patterns, severity mix, and major appearance/disappearance or rate/timing shifts. For each cited item, say what is directly visible and why it is relevant. Keep `Evidence` close to the log surface rather than turning it into a mini-diagnosis.
- `Reasoning` should explain what the stated evidence means and how it supports the summary. It may use domain knowledge and reasonable speculation, but it must stay tied to stated evidence from the current phase and may also refer explicitly to earlier phases when that materially clarifies the claim. Begin this subsection with exactly one line of the form `Certainty: highly likely`, `Certainty: moderately supported`, or `Certainty: speculative`. Use the certainty label that matches the weakest material inferential step in the subsection. Use `highly likely` when the conclusion is directly visible in the logs and follows from them without extra reasoning beyond what the logs explicitly show. Use `moderately supported` when the evidence is only partially visible or when ordinary domain knowledge or a bridging inference is needed. Use `speculative` when the evidence is indirect or incomplete and multiple plausible explanations remain. In the baseline, use this subsection to infer the normal system structure and behavior. In later phases, use it to explain what most likely changed operationally, how that affects the system, and what it says about the incident progression.

## Primary diagnosis
Use exactly these `###` subsections in this order:
- Summary
- Evidence
- Reasoning

For `## Primary diagnosis`:
- `Summary` should state only the core fault chain in plain terms, preferably in 1--2 sentences. Keep amplifiers, mitigation steps, and recovery detail out of this subsection unless they are part of the core diagnosis itself.
- `Evidence` should cite the main cross-phase evidence that supports that diagnosis. For each cited item, say what is directly visible and why it is relevant.
- `Reasoning` should explain why that evidence makes this diagnosis the best-supported interpretation. Put amplifiers, mitigation nuance, recovery detail, material alternatives, and unresolved ambiguities here when they matter, and begin with exactly one certainty line using the same scale as above. Use the certainty label that matches the weakest material inferential step in this subsection.

Additional requirements:
- When describing log templates, replace variable parts with "..."
- If the same observable request-flow signature appears again later, reference the earlier description and describe only the observable differences
- Use qualitative language for rate and latency changes unless "at least 2x" is warranted
- Write like an engineer producing a handoff-quality investigation note
\end{lstlisting}

\subsection{Example reference answer excerpt}
\label{app:example-answer}

\begin{lstlisting}[style=appendixtext]
## Incident synopsis
- Baseline -- Dashboard and Agent API operate normally: ELB routes to healthy Rails targets, health checks return 200, and PostgreSQL/PgBouncer show low contention.
- Trigger -- PostgreSQL saturates: connection timeouts and slow queries appear, dashboard `/healthcheck` fails with DB ping timeouts, ELB drains targets, and user requests flip to widespread 503s.
- Progression 1 -- Auto Scaling churn accelerates and recovery fails: new instances fail bootstrap (can't fetch deploy revision) and many health checks now fail with a "missing shard" error.
- Progression 2 -- Partial mitigation yields partial service: desired capacity is locked and instances are registered to the ELB, producing a mix of 200/500 health checks, fewer 503s, and some successful-but-slow logins.
- Progression 3 -- Scaling hits EC2 API limits: `RequestLimitExceeded` appears, successful launches/bootstraps drop, and the system remains degraded with continued dashboard errors.
- Progression 4 -- Major recovery actions applied: RDS is modified, AMI rollout and PgBouncer config change occur, DB/pgbouncer metrics shift to a "recovered" profile, health checks mostly succeed, but latency and intermittent errors persist.

## Reconstructed system
### Inferred purpose
A hosted CI service with a web "dashboard" (logins, build pages/log viewing, docs) served behind an AWS load balancer, plus a separate Agent API for build agents polling for jobs.

### Main services/components
- `aws-elb` (`elb-dashboard`): Front door/load balancer for the dashboard; emits access and health-check logs.
- `buildkite-web` (`dash-...`): Rails-like dashboard application instances; handles user endpoints and `/healthcheck`.
- `pgbouncer` (`pgb-1`): Connection pooler in front of the primary database; emits pool stats and connection failure warnings.
- `aws-rds-postgres` (`rds-primary`): PostgreSQL primary; emits periodic "metrics" snapshots plus connection timeout / slow-query warnings.
- `buildkite-agent-api` (`agent-...`): API for build agents; emits request logs and explicit retry markers.
- `aws-asg` (`asg-dashboard`): Auto Scaling activity for dashboard instances; emits launch/terminate plus bootstrap/EC2 API error markers.
- `monitoring` (`monitor-1`): Synthetic probes and PagerDuty events that detect/record unavailability.

### Main interactions
- Dashboard request path appears to be: client -> `aws-elb` -> `buildkite-web` -> (DB access via `pgbouncer`) -> `aws-rds-postgres` -> back to `aws-elb`, with the ELB also running periodic `/healthcheck` probes against each `buildkite-web` instance.
- The Agent API appears to talk directly to the same PostgreSQL (`aws-rds-postgres`) and will retry on failures.
- Auto Scaling launches/terminates dashboard instances and registers them to the ELB; instance bootstrapping appears to depend on fetching a deploy revision from a dashboard URL (inferred from bootstrap error logs referencing an HTTPS URL on the same domain).

## Timeline
[Phase-by-phase timeline omitted for space; the released benchmark file contains all phase sections.]

## Primary diagnosis
### Summary
PostgreSQL saturation caused the dashboard's DB-dependent `/healthcheck` to fail, leading the ELB to mark all dashboard targets unhealthy and return widespread 503s; recovery via Auto Scaling failed because new instances could not bootstrap (deploy revision fetch depended on the down site) and many ran an outdated health check referencing a missing DB shard.

### Evidence
- Cross-service health-check failure and ELB draining symptoms:
  - App: `"Healthcheck failed: database ping timeout waited_ms=... shard=two req=..."`
  - ELB: `"ELB healthcheck result target=... status=500 ..."`
  - ELB user impact: `"ELB request ... ... -> 503 no_healthy_backends ..."`
  The sequence shows health checks failing and user traffic failing due to no healthy backends.
- Database overload and connectivity failures coincide with health-check failures:
  - `"rds metrics ... profile=..."` plus frequent `"postgres connection timeout ..."` and `"slow query ..."`
  - PgBouncer elevated waiting: `"pgbouncer stats ... waiting=... avg_wait_ms=..."`
- Auto Scaling's inability to replace capacity is visible and tied to bootstrap/version issues:
  - `"asg launch instance=... ..."` followed by `"asg terminate instance=... reason="failed_elb_healthcheck""`
  - `"userdata fetch deploy revision failed url=https://buildkite.com/_secret/version err="...""`
  - Distinct app health-check failure: `"Healthcheck failed: database shard missing shard=one err="..." req=..."`
- Late improvements align with DB + bootstrap/pooling fixes:
  - `"rds modify instance_id=buildkite from=... to=... apply_immediately=true"`
  - `"ami rollout started ... revision_source=s3_file ..."`
  - `"pgbouncer config change pooling_mode=per_host ..."`
  followed by `"rds metrics ... profile=recovered"`, `"pgbouncer stats ... profile=recovered"`, and mostly-200 ELB health checks.

### Reasoning
Certainty: moderately supported  
The strongest, most direct chain in the logs is: DB overload (metrics/timeouts/slow queries) -> app `/healthcheck` explicitly failing due to DB ping timeout -> ELB recording 500 health-check results -> ELB returning 503 "no healthy backends" for user requests. That explains the initial outage state with only standard domain knowledge about how load balancer health checks gate routing. The prolonged inability to recover via scaling is best explained by the combination of (a) repeated launch/terminate cycles due to failed ELB health checks, (b) repeated bootstrap failures to fetch a deploy revision from a dashboard URL (a circular dependency under outage), and (c) a second, distinct health-check failure mode "database shard missing shard=one," which is most consistent with instances running mismatched/old code or configuration against a decommissioned shard. Alternatives like a pure ELB misconfiguration or a non-DB app crash are less consistent with the explicit DB timeout/shard-missing messages and the DB-side saturation warnings. The remaining uncertainty is the exact root cause of the DB saturation (e.g., traffic spike vs capacity reduction vs pooling change), because the logs show the saturation clearly but do not, by themselves, prove why the DB became overloaded--only that it did and that it dominated health-check and request behavior during the incident.
\end{lstlisting}

\section{LLM Judge Rubric}
\label{app:judge-rubric}

The judge receives the fixed question, the reference answer, and the predicted answer. The question defines the required report structure and certainty scale, while the reference answer defines the target incident story: the primary system, baseline behavior, trigger, progression, diagnosis, and uncertainty. The judge does not see the logs and is not allowed to use outside information.

The rubric scores seven axes using four labels: \emph{yes}, \emph{partial minor}, \emph{partial major}, and \emph{no}. \emph{Yes} means the axis matches cleanly. \emph{Partial minor} means the answer is mostly correct with only a local or non-central issue. \emph{Partial major} means there is a material but bounded defect. \emph{No} means the axis is absent or materially wrong. The axes separate report structure from incident semantics so that formatting errors are not confused with wrong diagnosis or chronology. Axis definitions are provided in Table~\ref{tab:judge-axes}.

The rubric allows useful extra specificity. Exact values, identifiers, hostnames, timestamps, counts, or log-template fragments are not penalized merely because the reference states the same point qualitatively. Such details are acceptable when they remain consistent with the reference incident story and are supported by the prediction's own observations. Conflicting or unsupported specificity is scored under support alignment; overconfidence about an unresolved trigger, dependency direction, or diagnosis is scored under uncertainty unless it changes the causal story itself. The judge records only the deepest independent mismatches and material extra claims, rather than counting one root error repeatedly.

The judge then assigns one of five ordered overall labels (Table~\ref{tab:judge-overall-verdicts}) by combining the per-axis scores based on instructions provided in the rubric. Diagnosis and timeline progression are the main story anchors. Support alignment and uncertainty determine whether the recovered story is properly supported and stated with the right level of confidence. System reconstruction provides context, and report structure determines whether the answer is a complete substitute for the reference.

For overall evaluation, we map the labels \emph{mismatch}, \emph{weak match}, \emph{partial match}, \emph{good match}, and \emph{strong match} to the numerical values \(0\), \(0.25\), \(0.50\), \(0.75\), and \(1.00\), respectively, as described in Sec.~\ref{sec:scoring}. When converting per-axis labels to per-axis scores, we map \emph{no}, \emph{partial major}, \emph{partial minor}, and \emph{yes} to 0.0, 0.33, 0.67, and 1.0, respectively.

The complete judge rubric is available in our code repository.

\begin{table}[h]
  \centering
  \small
  \caption{Evaluation axes used by the LLM judge.}
  \label{tab:judge-axes}
  \begin{tabular}{p{0.22\linewidth}p{0.72\linewidth}}
    \toprule
    Axis & Definition \\
    \midrule
    Structure & Compliance with the required report form: headings, subsections, certainty lines, and one synopsis bullet per reference phase. Semantic correctness is scored on the other axes. \\
    System reconstruction & Recovery of the same main system purpose, actors, and interactions. Extra plausible names or implementation details are allowed unless they change the main system, recenter a dependent component, or reverse a key dependency. \\
    Timeline structure & Agreement with the reference phase count, order, and time ranges. Small boundary drift and reference-preserving phase splits or merges are minor unless they hide an important state transition. \\
    Timeline progression & Recovery of the same baseline, trigger, major phase-to-phase changes, and final observable state. Missing or contradicting a major progression step is material. \\
    Diagnosis & Recovery of the same core fault chain. A more specific restatement is acceptable when it stays within the reference diagnosis and does not turn an unresolved mechanism into a required causal step. \\
    Support alignment & Whether the cited observations and reasoning support the same qualitative conclusions as the reference. This axis handles extra exactness, magnitude claims, local sequencing, and other log-surface details. \\
    Uncertainty & Match to the reference hedging on material uncertainty about the trigger, progression, a key dependency direction, or the main diagnosis. Mere log-surface precision is ignored on this axis. \\
    \bottomrule
  \end{tabular}
\end{table}

\begin{table}[t]
  \centering
  \small
  \caption{Overall verdicts used by the LLM judge.}
  \label{tab:judge-overall-verdicts}
  \begin{tabular}{p{0.20\linewidth}p{0.74\linewidth}}
    \toprule
    Verdict & Definition \\
    \midrule
    Strong match & A faithful substitute for the reference: the answer is semantically and structurally equivalent to the reference and covers all material content. \\
    Good match & A reliable recovery of the reference, but not a full substitute. The main incident story is the same, and any remaining semantic issues are minor and non-material. \\
    Partial match & A useful answer that gets the main causal story broadly right, but a material bounded defect or an accumulation of smaller semantic defects prevents using it as a reliable substitute. \\
    Weak match & The reference is still recognizable, but major defects dominate. The answer preserves fragments, observations, or broad phases, but does not cleanly recover the incident story. \\
    Mismatch & The prediction tells a materially different incident story and is not usable as a recovery of the reference. This includes switching to a different primary system, trigger, or fault chain. \\
    \bottomrule
  \end{tabular}
\end{table}

\clearpage

\section{Dataset Construction Cost}
\label{app:dataset_constr_cost}

Table~\ref{tab:pipeline-cost} shows the total benchmark token usage and cost breakdown. We use the following rates for the GPT-5.2 (high) model:
\begin{itemize}
    \item Input tokens: \$1.75 per million tokens;
    \item Cached input tokens: \$0.18 per million tokens;
    \item Output tokens: \$14 per million tokens.
\end{itemize}

We report token usage and costs for each stage of the pipeline, as well as a breakdown over the generator and verifier steps within each stage.

\begin{table}[h]
\centering
\caption{Token usage and cost breakdown across the three stages of the data generation pipeline.
Each stage reports the generator and the verifier usage and cost.
Cached tokens are a subset of the input tokens, and reasoning tokens are a subset of the output tokens.}
\label{tab:pipeline-cost}
\footnotesize
\setlength{\tabcolsep}{3.5pt}
\begin{tabular}{llrrrrrrrrr}
\toprule
& & & \multicolumn{4}{c}{Tokens} & \multicolumn{4}{c}{Cost [\$]} \\
\cmidrule(lr){4-7} \cmidrule(lr){8-11}
Stage & Category & Calls & Input & Cached & Output & Reasoning & Input & Cached & Output & Total \\
\midrule
\multirow{3}{*}{Incident model}
 & Generator       &   643 & 11{,}458{,}543 &  3{,}187{,}968 &  8{,}254{,}739 &  2{,}754{,}053 & 14.47 & 0.57 & 115.57 & 130.61 \\
 & Verifier & 2{,}730 & 44{,}687{,}360 & 18{,}951{,}040 & 14{,}351{,}580 & 13{,}511{,}415 & 45.04 & 3.41 & 200.92 & 249.37 \\
 & \textit{Subtotal} & \textit{3{,}373} & \textit{56{,}145{,}903} & \textit{22{,}139{,}008} & \textit{22{,}606{,}319} & \textit{16{,}265{,}468} & \textit{59.51} & \textit{3.99} & \textit{316.49} & \textit{379.99} \\
\midrule
\multirow{3}{*}{Incident simulator}
 & Generator       &   373 &  9{,}188{,}128 &  2{,}078{,}464 &  6{,}793{,}850 &  1{,}837{,}210 & 12.44 & 0.37 &  95.11 & 107.93 \\
 & Verifier & 1{,}855 & 47{,}917{,}225 & 19{,}464{,}960 & 12{,}079{,}385 & 11{,}704{,}445 & 49.79 & 3.50 & 169.11 & 222.41 \\
 & \textit{Subtotal} & \textit{2{,}228} & \textit{57{,}105{,}353} & \textit{21{,}543{,}424} & \textit{18{,}873{,}235} & \textit{13{,}541{,}655} & \textit{62.23} & \textit{3.88} & \textit{264.23} & \textit{330.34} \\
\midrule
\multirow{3}{*}{Reference answer}
 & Generator       &   180 &  3{,}167{,}550 &     877{,}184 &  1{,}044{,}137 &     103{,}753 &  4.01 & 0.16 &  14.62 &  18.78 \\
 & Verifier & 1{,}350 & 27{,}855{,}395 & 16{,}322{,}560 &  1{,}688{,}260 &  1{,}410{,}370 & 20.18 & 2.94 &  23.64 &  46.76 \\
 & \textit{Subtotal} & \textit{1{,}530} & \textit{31{,}022{,}945} & \textit{17{,}199{,}744} & \textit{2{,}732{,}397} & \textit{1{,}514{,}123} & \textit{24.19} & \textit{3.10} & \textit{38.25} & \textit{65.54} \\
\midrule
\multicolumn{2}{l}{\textbf{Pipeline total}} & \textbf{7{,}131} & \textbf{144{,}274{,}201} & \textbf{60{,}882{,}176} & \textbf{44{,}211{,}951} & \textbf{31{,}321{,}246} & \textbf{145.93} & \textbf{10.97} & \textbf{618.97} & \textbf{775.87} \\
\bottomrule
\end{tabular}
\end{table}

We note the asymmetry between generation and verification costs, with verification being around twice as expensive. The majority of the generation cost comes from feedback iterations where incorporation of every update requires the generator to completely re-generate the object with that update included. Improving the efficiency of feedback handling during generation (e.g., via in-place editing) could substantially reduce the generation costs, further increasing the asymmetry between generation and verification expenses.

\section{Evaluation Details}
\label{app:eval_details}

We use \texttt{LangChain} \citep{langchain2022harrison} to implement our ReAct-style \citep{yao2022react} agent that iteratively receives the input, reasons about the next action, and produces the next action as a tool call. The agent has access to a stateful Python interpreter that imports \texttt{pandas} \citep{mckinney2010data} and has access to all standard Python packages. This tool allows the agent to iteratively investigate the log data without rewriting the code in every new iteration.

We access all evaluated OpenAI models on Microsoft Azure; all other models (Anthropic and open-weight) are accessed via Amazon Bedrock. We use default sampling parameters for all models (temperature=1.0, top-p=1.0). We use a high (500k) output token limit per API request, and allow up to 200 tool calls per problem. For OpenAI models, we use default Azure input caching. Amazon Bedrock provides caching only for Anthropic models but not for the open-weight models. 

For all models, we use the following system prompt:

\begin{lstlisting}[style=appendixtext, basicstyle=\ttfamily\small, columns=fullflexible]
You are a log analysis agent. Your task is to answer a user's question about log data.

# Tools

python_notebook - executes Python in a stateful IPython kernel. State (variables, imports, DataFrames) is preserved across calls. pandas is pre-imported as pd.

# Log Data

A single CSV log file is available at {logs_path}. Load it with pd.read_csv(...) when needed.

# Rules

- Do NOT print large objects from python_notebook directly. Assign to variables, check sizes, and inspect in smaller parts to keep tool output concise.
- Do NOT create, modify, or delete any files.
- When you have the answer, respond with only the final answer text; no explanations or auxiliary text - just the complete final answer.
\end{lstlisting}

The user prompt consists of the question template shown in App.~\ref{app:fixed-question-and-example-answer}.

\section{Tool-call-budget sensitivity}
\label{app:tool-call-budget}

We test how the number of Python interpreter calls available to the agent affects scores and model rankings. We evaluate seven model configurations spanning the full-benchmark score range of 0.06--0.73 on a random sample of 30 \slsbench\ problems.

\paragraph{Setup.}
We use budgets of 0, 5, 10, 25, and 50 calls. The agent knows the total budget, receives the remaining budget with each tool output, and receives a \texttt{Budget reached} message when the budget is exhausted. Each tool output is capped at 12{,}000 characters and 1{,}000 lines to prevent oversized outputs at low budgets.

\paragraph{Scores.}
Table~\ref{tab:tool-budget-scores} reports mean scores for each budget. All models score zero without tool calls. Scores generally improve as the budget increases, and the model ordering is preserved across all nonzero budgets.

\begin{table}[htbp]
\centering
\caption{\textbf{\slsbench\ scores by tool-call budget.} Values are means over the 30-problem sample.}
\label{tab:tool-budget-scores}
\footnotesize
\setlength{\tabcolsep}{4pt}
\begin{tabular}{@{}llccccc@{}}
\toprule
Model & Effort & Budget 0 & Budget 5 & Budget 10 & Budget 25 & Budget 50 \\
\midrule
Nemotron 3 Nano & N/A & 0.00 & 0.02 & 0.04 & 0.06 & 0.07 \\
DeepSeek V3.2 & N/A & 0.00 & 0.09 & 0.17 & 0.20 & 0.29 \\
GPT-5.4 Nano & medium & 0.00 & 0.19 & 0.31 & 0.38 & 0.40 \\
Haiku 4.5 & N/A & 0.00 & 0.32 & 0.40 & 0.41 & 0.43 \\
Opus 4.6 & low & 0.00 & 0.44 & 0.45 & 0.50 & 0.50 \\
GPT-5.4 & low & 0.00 & 0.56 & 0.62 & 0.63 & 0.65 \\
GPT-5.4 & high & 0.00 & 0.58 & 0.64 & 0.70 & 0.71 \\
\bottomrule
\end{tabular}
\end{table}

\paragraph{Budget use.}
Table~\ref{tab:tool-budget-calls} reports the mean number of calls used per problem. With 50 calls available, DeepSeek V3.2 and GPT-5.4 Nano (medium) use about 37 calls on average, and Haiku 4.5 and Opus 4.6 (low) use about 16 and 11, respectively, indicating budget under-utilization.

\begin{table}[htbp]
\centering
\caption{\textbf{Tool calls used by budget.} Mean number of Python interpreter calls per problem.}
\label{tab:tool-budget-calls}
\footnotesize
\setlength{\tabcolsep}{4pt}
\begin{tabular}{@{}llccccc@{}}
\toprule
Model & Effort & Budget 0 & Budget 5 & Budget 10 & Budget 25 & Budget 50 \\
\midrule
Nemotron 3 Nano & N/A & -- & 4.23 & 6.57 & 12.73 & 19.53 \\
DeepSeek V3.2 & N/A & -- & 5.00 & 9.59 & 20.87 & 37.10 \\
GPT-5.4 Nano & medium & -- & 5.00 & 10.00 & 24.93 & 36.83 \\
Haiku 4.5 & N/A & -- & 4.07 & 7.30 & 13.70 & 15.50 \\
Opus 4.6 & low & -- & 4.23 & 6.93 & 10.40 & 10.97 \\
GPT-5.4 & low & -- & 5.00 & 9.73 & 20.00 & 22.37 \\
GPT-5.4 & high & -- & 5.00 & 9.83 & 23.43 & 35.52 \\
\bottomrule
\end{tabular}
\end{table}

\paragraph{Tool-output size.}
Table~\ref{tab:tool-budget-output} reports average output lengths in characters, with multipliers relative to the five-call budget. Per-call output size generally decreases as the budget grows, reaching 0.49--0.91 times the five-call value at a budget of 50. This pattern is consistent with concentrating log inspection into fewer, larger outputs when the budget is scarce.

\begin{table}[htbp]
\centering
\caption{\textbf{Tool-output size by budget.} Mean characters per tool output; parenthesized multipliers are relative to the five-call budget.}
\label{tab:tool-budget-output}
\footnotesize
\setlength{\tabcolsep}{3pt}
\begin{tabular}{@{}llccccc@{}}
\toprule
Model & Effort & Budget 0 & Budget 5 & Budget 10 & Budget 25 & Budget 50 \\
\midrule
Nemotron 3 Nano & N/A & -- & 957 ($1.00\times$) & 1020 ($1.07\times$) & 852 ($0.89\times$) & 845 ($0.88\times$) \\
DeepSeek V3.2 & N/A & -- & 1270 ($1.00\times$) & 1066 ($0.84\times$) & 728 ($0.57\times$) & 634 ($0.50\times$) \\
GPT-5.4 Nano & medium & -- & 1152 ($1.00\times$) & 1096 ($0.95\times$) & 1160 ($1.01\times$) & 1054 ($0.91\times$) \\
Haiku 4.5 & N/A & -- & 3743 ($1.00\times$) & 3669 ($0.98\times$) & 3104 ($0.83\times$) & 3257 ($0.87\times$) \\
Opus 4.6 & low & -- & 3887 ($1.00\times$) & 2688 ($0.69\times$) & 2333 ($0.60\times$) & 2335 ($0.60\times$) \\
GPT-5.4 & low & -- & 6629 ($1.00\times$) & 5577 ($0.84\times$) & 4126 ($0.62\times$) & 3904 ($0.59\times$) \\
GPT-5.4 & high & -- & 7301 ($1.00\times$) & 5904 ($0.81\times$) & 4193 ($0.57\times$) & 3570 ($0.49\times$) \\
\bottomrule
\end{tabular}
\end{table}

\section{Efficiency Analysis of Agent Trajectories}
\label{app:efficiency-analysis}

We reviewed ten GPT-5.4 (medium) trajectories and used a Python tool-output event as the unit of analysis. This unit matches the cost and latency mechanism of the agent: each output corresponds to a completed interpreter call whose result is passed back to the model. Counting the raw trajectories gives 361 such tool outputs in total.

\paragraph{Precomputable artifacts.}
We mark an output as precomputable when it is a log-derived artifact that can be produced before the agent commits to an incident hypothesis, phase boundary, selected time window, selected trace, or final diagnosis. These artifacts are the same kind of information an engineer usually asks for when first opening an unfamiliar log: schema and scale, timestamp range, service and host inventories, severity distributions, representative messages, global error and template summaries, and coarse temporal views. Under this definition, 75 of 361 (21\%) of the observed interpreter outputs  are precomputable. The trajectory-level counts are shown in Table~\ref{tab:trajectory-efficiency-counts}.

The agent is repeatedly computing generic summaries of the raw log. Instead, they could be pre-computed once per log file and provided as context before the agent begins the open-ended investigation. The agent can then spend its first reasoning steps interpreting the precomputed views rather than discovering the file format and rebuilding common summaries.

\paragraph{Orientation-and-planning prefix.}
For the parallelism analysis, we define the orientation prefix as the initial contiguous segment of tool outputs used to establish the shared investigation state before the main evidence-collection fan-out. This segment includes generic log orientation artifacts, such as schema, service inventory, message vocabulary, severity mix, templates, and coarse temporal views, together with early anchor-finding calls that identify phase boundaries, first anomalies, or candidate causal components. The prefix ends immediately when the agent begins executing downstream tasks in order to substantiate, quantify, or cross-check the incident explanation.

With this definition, the orientation-and-planning prefix accounts for 108 of 361 outputs (29.9\%). The prefix is not identical to the precomputable set: some precomputable summaries appear later, and some early anchor-finding calls are incident-specific rather than generic.

After this prefix, the dominant pattern is fan-out. The agent typically has enough information to separate analyses such as service-level timelines, non-INFO template breakdowns, phase-level statistics, trace or request examples, support-plane event timelines, and recovery checks. These streams use the same log messages as input but do not depend on each other's intermediate outputs. A parallel implementation could issue these tasks concurrently, then aggregate the results for the final report synthesis.

\begin{table}[t]
\centering
\caption{\textbf{Trajectory-level efficiency counts.} Tool outputs are Python interpreter outputs in the raw trajectory. Precomputable outputs are generic log artifacts available before agent reasoning. Prefix outputs are the initial orientation-and-planning calls before the main evidence-collection fan-out.}
\label{tab:trajectory-efficiency-counts}
\footnotesize
\setlength{\tabcolsep}{5pt}
\begin{tabular}{@{}l l r r r r@{}}
\toprule
Trajectory & Problem ID & Tool outputs & Precomputable & Prefix & Prefix fraction \\
\midrule
T1  & \texttt{03a9b743} & 39 & 7  & 10 & 25.6\% \\
T2  & \texttt{1cddedc3} & 36 & 10 & 10 & 27.8\% \\
T3  & \texttt{0ff91f6a} & 35 & 7  & 9  & 25.7\% \\
T4  & \texttt{0a7c69ff} & 42 & 8  & 11 & 26.2\% \\
T5  & \texttt{2d21ffed} & 32 & 7  & 11 & 34.4\% \\
T6  & \texttt{0b9e6bdf} & 39 & 7  & 11 & 28.2\% \\
T7  & \texttt{2ee83142} & 38 & 5  & 12 & 31.6\% \\
T8  & \texttt{1c25d7c3} & 35 & 6  & 13 & 37.1\% \\
T9  & \texttt{0a0114df} & 33 & 9  & 13 & 39.4\% \\
T10 & \texttt{2d3c2d7e} & 32 & 9  & 8  & 25.0\% \\
\midrule
Total & -- & 361 & 75 & 108 & -- \\
\bottomrule
\end{tabular}
\end{table}

\paragraph{Trajectory parallelization example.}
The bottom panel of Fig.~\ref{fig:efficiency_analysis} uses T9 (\texttt{0a0114df}) as a concrete example of the fan-out pattern. The reconstruction is code-level rather than merely narrative: outputs 1--13 form the shared orientation-and-planning prefix, and each later Python output is assigned to a stream only when it can be run from that shared state without consuming another post-prefix stream's intermediate result. The shared state consists of the parsed log messages, service-specific views (\texttt{edge-gw}, \texttt{rtm}, \texttt{app-api}, \texttt{db}, and \texttt{monitor}), extracted fields such as \texttt{trace\_id}, \texttt{rtm\_target}, HTTP status, error kind, and DB counters, plus the three incident anchors: the \texttt{rtm-3} crash/failure window around \texttt{12:20}, DB/authentication degradation around \texttt{12:27}, and the mitigation boundary around \texttt{12:34}. Table~\ref{tab:t9-parallel-workstreams} expands the four stream families shown in the figure.

\begin{table}[t]
\centering
\caption{\textbf{Code-level construction of Fig.~\ref{fig:efficiency_analysis} (bottom) for T9.} Calls are Python tool-output events. The bottom four rows match the stream labels in the figure and sum to the 20 outputs after the 13-call orientation prefix.}
\label{tab:t9-parallel-workstreams}
\footnotesize
\setlength{\tabcolsep}{3.5pt}
\renewcommand{\arraystretch}{1.24}
\begin{tabular}{@{}>{\raggedright\arraybackslash}p{0.17\linewidth}c >{\raggedright\arraybackslash}p{0.31\linewidth}>{\raggedright\arraybackslash}p{0.41\linewidth}@{}}
\toprule
Stage / stream & Calls & Code functionality & Concrete role in summary construction \\
\midrule
Orientation and planning & 13 & Load and parse the dataframe; compute schema, level/service/template summaries, coarse per-minute views, and short anomaly windows; build reusable service slices and phase anchors. & Establishes the shared state used by every stream: \texttt{rtm-3} websocket failures and a \texttt{SIGSEGV} crash near \texttt{12:20}; later session-validation/history failures tied to DB pressure near \texttt{12:27}; and mitigation/recovery near \texttt{12:34}. \\
\midrule
Failure-chain traces & 8 & Select anomalous request families---failed websocket upgrades, rejected upgrades, and history warnings---then read all rows with the same \texttt{trace\_id}. & Produces request-level chains: failed upgrades target \texttt{rtm-3}; rejected upgrades pass through \texttt{session validate timeout} with \texttt{db\_err=pool\_exhausted}; history warnings line up with app-api timeouts and gateway 503/504 responses. \\
Service timelines & 3 & Group service-specific slices by minute or phase using parsed fields such as \texttt{rtm\_target}, HTTP status, DB counters, and monitor target. & Shows component evolution: in \texttt{[12:20,12:27)}, \texttt{rtm-3} has 420 failed upgrades and 0 successful upgrades while other RTM targets continue serving; later timelines expose DB/app-api degradation. \\
Before/after traces & 4 & Sample representative healthy traces before the incident and recovery traces after mitigation, then read the complete trace rows. & Gives contrastive examples: normal history, websocket, and message-ingest paths before the crash, plus a successful \texttt{rtm-3} websocket handshake after the \texttt{12:34} mitigation boundary. \\
Per-phase statistics & 5 & Loop over phase windows and count levels, templates, HTTP statuses, websocket outcomes, API events, and DB event types. & Compresses the trajectory into phase evidence: baseline behavior, RTM-targeted trigger, DB/auth/API degradation, and partial recovery. \\
\bottomrule
\end{tabular}
\end{table}

The resulting dependency graph is a fan-out/fan-in computation. Once the orientation state is fixed, target accounting does not need trace reconstruction, trace reconstruction does not need phase rollups, and recovery checks do not need DB timeline summaries. A parallel implementation can therefore issue the four stream families concurrently and aggregate their outputs before final report synthesis, without changing the evidence used by the serialized trajectory.

\section{Benchmark and Evaluation Validity}
\label{app:benchmark-validity}

\subsection{Judge robustness}
\label{app:judge-robustness}

To ensure our LLM judges grade responses based on substance and meaning, rather than superficial adherence to the ground-truth section structure or wording, we test the judges from Sec.~\ref{sec:scoring} on reference summaries modified to preserve or alter their meaning. Each evaluation pairs an original reference with its modified version as the prediction. We measure lexical overlap using word-based Jaccard similarity.

\paragraph{Meaning-preserving changes.}
We use Opus 4.6 to rephrase all 90 reference summaries. The model is instructed to preserve the meaning, substance, and main report sections required by the benchmark question while changing wording and internal structure as much as possible. We manually verify ten rephrasings for meaning preservation and substantial rewording. The average word-Jaccard similarity between the original summaries and their reworded versions is 0.56, indicating low overlap.

\paragraph{Meaning-altering changes.}
We also generate LLM-modified summaries that change the meaning and substance of the ground-truth summaries, while preserving their structure and lexical content as much as possible. Their average word-Jaccard similarity is 0.87, indicating high overlap. Similarly, we manually inspect ten modified versions to ensure adequate task execution.

\paragraph{Results.}
Table~\ref{tab:judge-robustness} reports the scores for both experiments. Meaning-preserving rephrasings (same meaning, different structure) receive an average score of 0.99, and meaning-altering changes (different meaning, same structure) receive 0.17. Both judges retain near-maximal scores under rephrasing and strongly penalize substantive changes. These experiments suggest that our LLM judges grade responses based on their meaning and substance rather than superficial structure or lexical matching.

\begin{table}[htbp]
\centering
\caption{\textbf{Judge robustness to reference modifications.} Word-Jaccard is the average lexical similarity between original and modified summaries. Judge scores use GPT-5.2 (medium) and Sonnet 4.6 (low); Average is their mean, as used for \slsbench\ scoring.}
\label{tab:judge-robustness}
\footnotesize
\begin{tabular}{@{}lcccc@{}}
\toprule
Modification & Word-Jaccard & GPT-5.2 & Sonnet 4.6 & Average \\
\midrule
Meaning-preserving & 0.56 & 0.98 & 0.99 & 0.99 \\
Meaning-altering   & 0.87 & 0.14 & 0.20 & 0.17 \\
\bottomrule
\end{tabular}
\end{table}

\subsection{Human evaluation}
\label{app:human-evaluation}

LLM judges grade predicted summaries by comparing them to the reference summaries only, without involving the log streams, which helps to make evaluation simpler, and reduce its time and cost. Here, we test if this evaluation approach correlates with how human evaluators would grade the predicted summaries against the actual log streams from which they were produced.

We compare LLM judge scores with human assessments of predicted summaries against the logs. We select five models closest to five equally spaced scores across the full-benchmark score range of 0.06--0.73. We randomly sample five \slsbench\ problems and use the same problems for every model, yielding 25 predicted summaries.

\paragraph{Evaluation procedure.}
One human evaluator scores the summaries using a condensed, rule-based version of the LLM judge rubric (App.~\ref{app:judge-rubric}). The condensed rubric retains the core evaluation criteria and omits fine-grained requirements to reduce evaluation time. The evaluator also receives the incident model as a concise overview of the log stream's structure to accelerate grading. All assessments are required to be based on the corresponding log streams. Model identities are unknown to the evaluator.

\paragraph{Results and scope.}
Table~\ref{tab:human-evaluation} reports mean scores over the five shared problems. The combined \slsbench\ scores and human scores produce identical model rankings. Higher human scores for the strongest models are consistent with the condensed rubric's greater leniency. The study is limited to one evaluator and five problems, with a condensed rubric and access to the incident model. Use of the condensed rubric by the human evaluator also limits direct comparison of the absolute scores. However, the results broadly indicate that our evaluation approach aligns with human grading of the predicted summaries based on log streams.

\begin{table}[htbp]
\centering
\caption{\textbf{LLM judge and human evaluation scores.} Both columns are means over the same \textbf{five} sampled \slsbench\ problems. The \slsbench\ score averages the two LLM judges.}
\label{tab:human-evaluation}
\footnotesize
\begin{tabular}{@{}lcc@{}}
\toprule
Model & \slsbench\ score & Human score \\
\midrule
Nemotron 3 Nano        & 0.00 & 0.10 \\
DeepSeek V3.2          & 0.10 & 0.20 \\
GPT-5.4 Nano (medium)  & 0.38 & 0.35 \\
Opus 4.6 (low)         & 0.53 & 0.65 \\
GPT-5.4 (high)         & 0.63 & 0.95 \\
\bottomrule
\end{tabular}
\end{table}

\subsection{Comparison with external benchmarks}
\label{app:external-benchmarks}

We compare \slsbench\ model rankings with results on o11y-bench and OpenRCA to assess their agreement across independent observability tasks.

\renewcommand\thefootnote{\arabic{footnote}}
\setcounter{footnote}{0}

\paragraph{o11y-bench.}
Grafana Labs' o11y-bench\footnote{\url{https://github.com/grafana/o11y-bench}} evaluates agents on observability and site reliability engineering workflows in a live Grafana stack, including querying metrics, logs, and traces and investigating incidents. An LLM judge grades tasks using task-specific rubrics. The $\mathrm{Pass}^{3}$ score is the fraction of tasks solved in all three runs. We use published leaderboard scores for OpenAI and Anthropic models and evaluate the open-weight models ourselves.

Table~\ref{tab:external-o11y} compares scores for 15 model configurations. The overall Spearman rank correlation is 0.81, with correlations of 0.83 for OpenAI, 1.00 for Anthropic, and 0.90 for open-weight models. A notable ranking difference is GPT-5.4 Nano (high), which scores above both GPT-5.4 Mini configurations on o11y-bench. The Anthropic ordering is preserved exactly.

\begin{table}[htbp]
\centering
\caption{\textbf{\slsbench\ and o11y-bench scores.} A superscript $*$ marks o11y-bench scores from the published leaderboard; unmarked o11y-bench scores are from our runs.}
\label{tab:external-o11y}
\footnotesize
\begin{tabular}{@{}lcc@{}}
\toprule
Model & \slsbench\ score & o11y-bench $\mathrm{Pass}^{3}$ \\
\midrule
\multicolumn{3}{@{}l}{\textit{OpenAI}} \\
GPT-5.4 (high)          & 0.73 & $0.63^{*}$ \\
GPT-5.4 (low)           & 0.67 & $0.56^{*}$ \\
GPT-5.4 Mini (high)     & 0.61 & $0.29^{*}$ \\
GPT-5.4 Mini (low)      & 0.48 & $0.21^{*}$ \\
GPT-5.4 Nano (high)     & 0.44 & $0.48^{*}$ \\
GPT-5.4 Nano (low)      & 0.28 & $0.19^{*}$ \\
\addlinespace[3pt]
\multicolumn{3}{@{}l}{\textit{Anthropic}} \\
Opus 4.6 (high)        & 0.61 & $0.59^{*}$ \\
Sonnet 4.6 (low)       & 0.58 & $0.51^{*}$ \\
Opus 4.6 (low)         & 0.53 & $0.49^{*}$ \\
Haiku 4.5 (off)        & 0.46 & $0.48^{*}$ \\
\addlinespace[3pt]
\multicolumn{3}{@{}l}{\textit{Open--Weight}} \\
GLM-5 (medium)         & 0.49 & 0.38 \\
MiniMax-M2 (medium)    & 0.29 & 0.29 \\
DeepSeek V3.2 (medium) & 0.27 & 0.32 \\
Ministral 3 (medium)   & 0.12 & 0.13 \\
Nemotron 3 Nano (medium) & 0.06 & 0.06 \\
\bottomrule
\end{tabular}
\end{table}

\paragraph{OpenRCA.}
OpenRCA \citep{xu2025openrca} evaluates root-cause localization from real production telemetry, scoring categorical outputs for failure time, component, and reason. We use its Telemetry subset of 51 problems and select five models spread across the \slsbench\ score range. GPT-5.4, GPT-5.4 Nano, and Sonnet 4.6 use low reasoning effort to limit evaluation time.

Table~\ref{tab:external-openrca} reports OpenRCA's Correct and Partial scores. The Partial scores have a Spearman rank correlation of 0.90 with \slsbench. The main ranking difference is Haiku 4.5, which scores above GPT-5.4 Nano (low) on \slsbench\ and below it on both OpenRCA metrics.

\begin{table}[htbp]
\centering
\caption{\textbf{\slsbench\ and OpenRCA scores.} OpenRCA results cover the 51-problem Telemetry subset; \slsbench\ scores are from the full benchmark.}
\label{tab:external-openrca}
\footnotesize
\begin{tabular}{@{}lccc@{}}
\toprule
Model & \slsbench\ score & RCA Correct & RCA Partial \\
\midrule
GPT-5.4 (low)      & 0.67 & 0.14 & 0.22 \\
Sonnet 4.6 (low)   & 0.58 & 0.14 & 0.21 \\
Haiku 4.5          & 0.46 & 0.04 & 0.08 \\
GPT-5.4 Nano (low) & 0.28 & 0.12 & 0.20 \\
Nemotron 3 Nano    & 0.06 & 0.02 & 0.07 \\
\bottomrule
\end{tabular}
\end{table}

Because both benchmarks measure different tasks and grading criteria than \slsbench\ and because the OpenRCA evaluation covers only five models on 51 cases, we treat these as directional rank comparisons rather than absolute-score comparisons.

However, strong rank agreement with two independent real-world benchmarks provides evidence that \slsbench\ performance is predictive of agent performance on real data and on log analytics tasks that differ from \slsbench\ itself.

\section{Extra Figures}

\begin{figure}[h]
    \centering
    \includegraphics[width=0.75\linewidth]{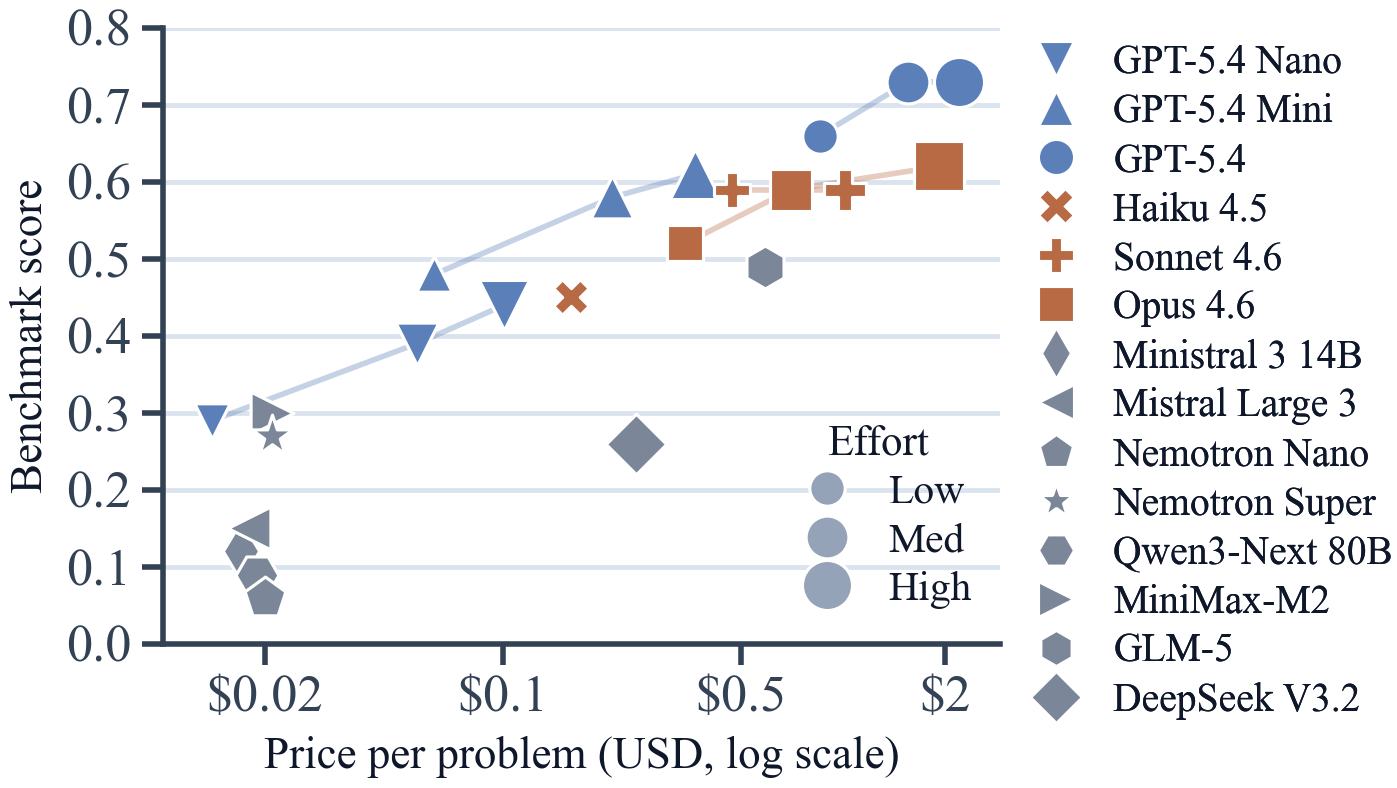}
    \caption{Benchmark score vs. average cost per problem for all evaluated models.}
    \label{fig:results_all_models}
\end{figure}

%%%%%%%%%%%%%%%%%%%%%%%%%%%%%%%%%%%%%%%%%%%%%%%%%%%%%%%%%%%%

\newpage
\section*{NeurIPS Paper Checklist}

\begin{enumerate}

\item {\bf Claims}
    \item[] Question: Do the main claims made in the abstract and introduction accurately reflect the paper's contributions and scope?
    \item[] Answer: \answerYes{} % Replace by \answerYes{}, \answerNo{}, or \answerNA{}.
    \item[] Justification: Benchmark and data generation pipeline are presented in Section \ref{sec:benchmark}. Evaluations are presented in Section \ref{sec:results}.

\item {\bf Limitations}
    \item[] Question: Does the paper discuss the limitations of the work performed by the authors?
    \item[] Answer: \answerYes{} % Replace by \answerYes{}, \answerNo{}, or \answerNA{}.
    \item[] Justification: We discuss limitations in Section \ref{sec:benchmark} and Section \ref{sec:conclusion}.

\item {\bf Theory assumptions and proofs}
    \item[] Question: For each theoretical result, does the paper provide the full set of assumptions and a complete (and correct) proof?
    \item[] Answer: \answerNA{} % Replace by \answerYes{}, \answerNo{}, or \answerNA{}.
    \item[] Justification: The paper does not include theoretical results.

    \item {\bf Experimental result reproducibility}
    \item[] Question: Does the paper fully disclose all the information needed to reproduce the main experimental results of the paper to the extent that it affects the main claims and/or conclusions of the paper (regardless of whether the code and data are provided or not)?
    \item[] Answer: \answerYes{} % Replace by \answerYes{}, \answerNo{}, or \answerNA{}.
    \item[] Justification: We provide all information, data, and code to reproduce the main results. Our released code contains all necessary information to run the generation and evaluation pipelines. Also, our released data can be directly used with the code to reproduce the main evaluation results.

\item {\bf Open access to data and code}
    \item[] Question: Does the paper provide open access to the data and code, with sufficient instructions to faithfully reproduce the main experimental results, as described in supplemental material?
    \item[] Answer: \answerYes{} % Replace by \answerYes{}, \answerNo{}, or \answerNA{}.
    \item[] Justification: For double-blind review purposes, our submission provides links to anonymized data and code repositories containing necessary information to use them. For the public version of the paper we will provide de-anonymized links to our data and code repositories directly in the paper.

\item {\bf Experimental setting/details}
    \item[] Question: Does the paper specify all the training and test details (e.g., data splits, hyperparameters, how they were chosen, type of optimizer) necessary to understand the results?
    \item[] Answer: \answerYes{} % Replace by \answerYes{}, \answerNo{}, or \answerNA{}.
    \item[] Justification: Our benchmark has only one split that was used to obtain the reported results. The LLM providers and generation parameters are described in the Appendix.

\item {\bf Experiment statistical significance}
    \item[] Question: Does the paper report error bars suitably and correctly defined or other appropriate information about the statistical significance of the experiments?
    \item[] Answer: \answerYes{} % Replace by \answerYes{}, \answerNo{}, or \answerNA{}.
    \item[] Justification: Table \ref{tab:model_cost_perf} provides standard error of the mean for the score, and standard deviation for the agent runtime.

\item {\bf Experiments compute resources}
    \item[] Question: For each experiment, does the paper provide sufficient information on the computer resources (type of compute workers, memory, time of execution) needed to reproduce the experiments?
    \item[] Answer: \answerYes{} % Replace by \answerYes{}, \answerNo{}, or \answerNA{}.
    \item[] Justification: We provide the dataset construction costs in Table~\ref{tab:pipeline-cost} as well as the cost per problem for each evaluated model in Table~\ref{tab:model_cost_perf}.
    
\item {\bf Code of ethics}
    \item[] Question: Does the research conducted in the paper conform, in every respect, with the NeurIPS Code of Ethics \url{https://neurips.cc/public/EthicsGuidelines}?
    \item[] Answer: \answerYes{} % Replace by \answerYes{}, \answerNo{}, or \answerNA{}.
    \item[] Justification: Our research adheres to the Code of Ethics.

\item {\bf Broader impacts}
    \item[] Question: Does the paper discuss both potential positive societal impacts and negative societal impacts of the work performed?
    \item[] Answer: \answerYes{} % Replace by \answerYes{}, \answerNo{}, or \answerNA{}.
    \item[] Justification: We discuss positive impact of our research in the Introduction. We do not see immediate negative societal impact of our research.
    
\item {\bf Safeguards}
    \item[] Question: Does the paper describe safeguards that have been put in place for responsible release of data or models that have a high risk for misuse (e.g., pre-trained language models, image generators, or scraped datasets)?
    \item[] Answer: \answerNA{} % Replace by \answerYes{}, \answerNo{}, or \answerNA{}.
    \item[] Justification: Our presented work does not have a high risk of misuse.

\item {\bf Licenses for existing assets}
    \item[] Question: Are the creators or original owners of assets (e.g., code, data, models), used in the paper, properly credited and are the license and terms of use explicitly mentioned and properly respected?
    \item[] Answer: \answerYes{} % Replace by \answerYes{}, \answerNo{}, or \answerNA{}.
    \item[] Justification: Our benchmark consists of synthetic data grounded in publicly available real-world incident reports; we provide URLs to these reports in our code repository and Croissant metadata.

\item {\bf New assets}
    \item[] Question: Are new assets introduced in the paper well documented and is the documentation provided alongside the assets?
    \item[] Answer: \answerYes{} % Replace by \answerYes{}, \answerNo{}, or \answerNA{}.
    \item[] Justification: We provide all necessary documentation in this paper and the released code and data repositories.

\item {\bf Crowdsourcing and research with human subjects}
    \item[] Question: For crowdsourcing experiments and research with human subjects, does the paper include the full text of instructions given to participants and screenshots, if applicable, as well as details about compensation (if any)? 
    \item[] Answer: \answerNA{} % Replace by \answerYes{}, \answerNo{}, or \answerNA{}.
    \item[] Justification: The paper does not involve crowdsourcing nor research with human subjects.

\item {\bf Institutional review board (IRB) approvals or equivalent for research with human subjects}
    \item[] Question: Does the paper describe potential risks incurred by study participants, whether such risks were disclosed to the subjects, and whether Institutional Review Board (IRB) approvals (or an equivalent approval/review based on the requirements of your country or institution) were obtained?
    \item[] Answer: \answerNA{} % Replace by \answerYes{}, \answerNo{}, or \answerNA{}.
    \item[] Justification: The paper does not involve crowdsourcing nor research with human subjects.

\item {\bf Declaration of LLM usage}
    \item[] Question: Does the paper describe the usage of LLMs if it is an important, original, or non-standard component of the core methods in this research? Note that if the LLM is used only for writing, editing, or formatting purposes and does \emph{not} impact the core methodology, scientific rigor, or originality of the research, declaration is not required.
    %this research? 
    \item[] Answer: \answerYes{} % Replace by \answerYes{}, \answerNo{}, or \answerNA{}.
    \item[] Justification: We use LLMs as integral parts of our synthetic data generation pipeline (generators and verifiers). We also use LLM judges to evaluate the predicted answers against the ground truth answers. We provide details about the usage of the LLMs in Section~\ref{sec:benchmark} and Appendix. We also open-source all the code and prompts used for data generation, validation, and answer scoring. 

\end{enumerate}

\end{document}